\documentclass[acmsmall,screen]{acmart}
\usepackage{graphicx}
\usepackage{amsfonts}
\usepackage{hyperref}
\usepackage{url}
\usepackage{booktabs}
\usepackage{nicefrac}
\usepackage{microtype}
\usepackage{comment}
\usepackage{lipsum}
\graphicspath{ {./} }
\usepackage[ruled,vlined,linesnumbered]{algorithm2e}
\usepackage{mathtools}
\usepackage{xcolor}
\usepackage{array}
\usepackage{cleveref}
\usepackage{enumitem}
\usepackage{wrapfig}
\usepackage{booktabs, tabularx}
\usepackage{mathtools,trimclip,stackengine,scalerel}
\usepackage{eucal}
\usepackage{caption}
\graphicspath{ {./images/} }
\usepackage{ifthen}
\usepackage{amsmath}
\usepackage{comment}
\usepackage{thmtools, thm-restate}
\usepackage{hyperref}
\usepackage{longtable}
\usepackage{ifthen}
\usepackage{environ}
\newtheorem{problem}{Problem}
\newboolean{is_conference}
\setboolean{is_conference}{false} 

\newcommand{\tool}{VHAGaR\xspace}
\newcommand{\Dana}[1]{\textcolor{purple}{\bf Dana: #1}}

\begin{document}

\title{Verification of Neural Networks’ Global Robustness}

\author{Anan Kabaha}
\orcid{0000-0002-0969-6169}
\affiliation{%
  \institution{Technion}
  \city{Haifa}
  \country{Israel}
}
\email{anan.kabaha@campus.technion.ac.il}

\author{Dana Drachsler Cohen}
\orcid{0000-0001-6644-5377}
\affiliation{%
  \institution{Technion}
  \city{Haifa}
  \country{Israel}
}
\email{ddana@ee.technion.ac.il}



\begin{abstract}
Neural networks are successful in various applications but are also susceptible to adversarial attacks.
To show the safety of network classifiers, many verifiers have been introduced to reason about the local robustness of a given input to a given perturbation. While successful, local robustness cannot generalize to unseen inputs. Several works analyze global robustness properties, however, neither can provide a precise guarantee about the cases where a network classifier does not change its classification.   
In this work, we propose a new global robustness property for classifiers aiming at finding the \emph{minimal globally robust bound}, which naturally extends the popular local robustness property for classifiers. We introduce \tool, an anytime verifier for computing this bound.
\tool relies on three main ideas: encoding the problem as a mixed-integer programming and pruning the search space by identifying dependencies stemming from the perturbation or the network's computation and generalizing adversarial attacks to unknown inputs.
 We evaluate \tool on several datasets and classifiers and show that, given a three hour timeout, the average gap between the lower and upper bound on the minimal globally robust bound computed by \tool is 1.9, while the gap of an existing global robustness verifier is 154.7. Moreover, \tool is 130.6x faster than this verifier.
 Our results further indicate that leveraging dependencies and adversarial attacks makes \tool 78.6x faster.
\end{abstract}

\maketitle
\section{Introduction}
\label{sec:introduction}

Deep neural networks are successful in various tasks but are also susceptible to adversarial examples: malicious input perturbations designed to deceive the network. Many adversarial attacks target image classifiers and compute either an imperceptible change, bounded by a small $\epsilon$ with respect to an $L_{p}$ norm (e.g., $p=0,1,2,\infty$)~\citep{g_ref_3,g_ref_4,g_ref_5,g_ref_6,g_ref_8,g_ref_9,g_ref_70},
or a perceivable change obtained by perturbing a semantic feature that does not change the semantics of the input, such as brightness, translation, or rotation~\citep{g_ref_12,g_ref_14,g_ref_15,g_ref_17,g_ref_18}.

A popular safety property for understanding a network's robustness to such attacks is \emph{local robustness}.
Local robustness is parameterized by an input and its neighborhood. For a network classifier, the goal is to prove that the network classifies all inputs in the neighborhood the same.
Several local robustness verifiers have been introduced for checking robustness in an $\epsilon$-ball~\citep{g_ref_33,g_ref_26,g_ref_28,g_ref_20,g_ref_21,g_ref_23} or 
a feature neighborhood~\citep{g_ref_37,g_ref_13}. However, local robustness is limited to reasoning about a single neighborhood at a time. Thus, the network designer has to reason separately about every input that may arise in practice. This raises several issues. First, the space of inputs typically has a high dimensionality, making it impractical to be covered by a finite set of neighborhoods. 
Second, the robustness of a set of neighborhoods does not imply that unseen neighborhoods are robust. 
Third, even if the network designer can identify a finite set of relevant neighborhoods, existing local robustness verifiers take non-negligible time to reason about a single neighborhood. Thus, running them on a very large number of neighborhoods is impractical.  
Hence, local robustness does not enable to fully understand the robustness level of a network to a given perturbation for all inputs.

These issues gave rise to reasoning about a network's robustness over all possible inputs, known as \emph{global robustness}. Several works analyze global robustness properties. They can be categorized into two primary groups: precise analysis~\citep{g_ref_30,g_ref_31,g_ref_1,g_ref_2} 
and sampling-based analysis~\citep{g_ref_41,g_ref_42,g_ref_43,g_ref_44}. 
Existing precise analysis techniques focus primarily on the $L_\infty$ $\epsilon$-ball and a global robustness property defined over the network's stability~\citep{g_ref_1,g_ref_2,g_ref_30}. The network's stability 
is the maximum difference of the output vectors of an input and a perturbed example in that input’s $\epsilon$-ball. 
This approach cannot capture the desired robustness property for classifiers, which is that the inputs' classification does not change under a given perturbation. 
 The second kind of analyzers computes a probabilistic global robustness bound by analyzing the local robustness of samples from the input space~\citep{g_ref_43,g_ref_44} or a given dataset~\citep{g_ref_41,g_ref_42}. 
 Sampling-based analyzers scale better than the precise analyzers but only provide a lower bound on the actual global robustness bound. 

In this work, we propose a new global robustness property designated for classifiers, which is general to any perturbation such as the $L_\infty$ perturbation or feature perturbations. 
Intuitively, a classifier is globally robust to a given perturbation if for any input for which the network is confident enough in its classification, this perturbation does not cause the network to change its classification. 
We focus on such inputs because it is inevitable that inputs near the decision boundaries, where the network's classification confidence is low, are misclassified when adding perturbations. 
Note that, like previous works~\cite{g_ref_30,g_ref_31,g_ref_1,g_ref_2}, our global robustness property considers \emph{any} input, including meaningful inputs and meaningless noise. In a scenario where the focus is only on meaningful inputs, our property can theoretically be restricted to this space. However, this requires a formal characterization of the meaningful input space, which is generally nonexistent. For such scenario, our property, which considers any input, can be viewed as over-approximating the space of meaningful inputs for which the network is confident enough. 

We address the problem of computing the \emph{minimal globally robust bound} of a given network and perturbation. Namely, any input that the network classifies with a confidence that is at least this bound is not misclassified under this perturbation.
Although our global robustness property is not the first to consider any input, it is the first to consider the classifier's minimal bound satisfying global robustness. This difference is similar to the difference between verifying local robustness at a \emph{given} $\epsilon$-ball 
and computing the \emph{maximal} $\epsilon$-ball that is locally robust. 
Our problem is highly challenging for two main reasons. First, 
it is challenging to determine for a \emph{given} confidence whether a classifier is globally robust since it requires to determine the robustness of a very large space of unknown inputs; 
 even local robustness verification, reasoning about a single input's neighborhood, is under active research.
Second, our problem requires to compute the \emph{minimal} globally robust bound, and thus it requires reasoning about the global robustness over a large number of confidences.   

\begin{figure*}[t]
    \centering
  \includegraphics[width=1\linewidth, trim=0 390 30 0, clip,page=4]{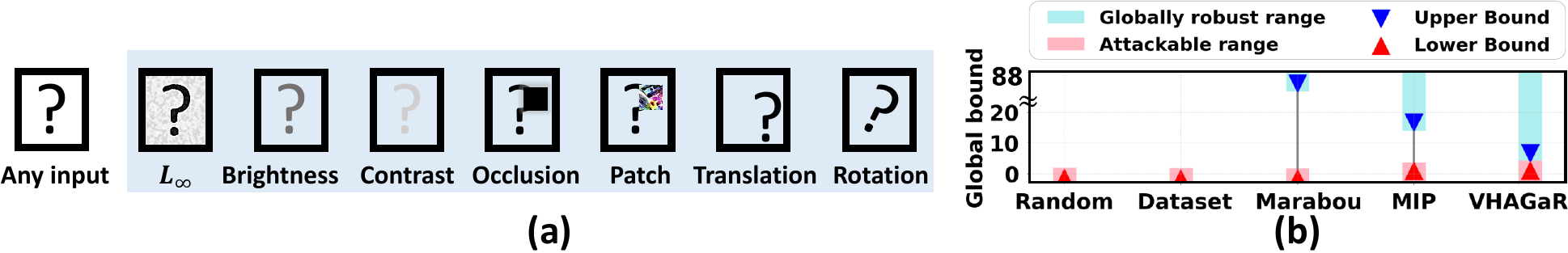}
    \caption{(a)~The perturbations supported by \tool. (b)~\tool's upper and lower bounds on the minimal globally robust bound vs. sampling approaches, Marabou and a MIP-only variant of \tool.}
    \label{fig::intro}
\end{figure*}

We introduce \tool
, a \textbf{V}erifier of \textbf{H}azardous \textbf{A}ttacks for proving \textbf{G}lob\textbf{a}l \textbf{R}obustness that computes the minimal globally robust bound of a network classifier, under a given perturbation. 
\tool relies on three key ideas. First, it encodes the problem as a mixed-integer programming (MIP), enabling efficient optimization through existing MIP solvers. Further, \tool is designed as an anytime algorithm and asks the MIP solver to compute a lower and an upper bound on the minimal globally robust bound. 
Second, it prunes the search space by computing dependencies stemming from the perturbation or the network's computation. 
Third, it executes a \emph{hyper-adversarial attack}, generalizing adversarial attacks to unknown inputs, to efficiently compute suboptimal lower bounds on the robustness bound that further prune the search space. 
Our attack also identifies \emph{optimization hints}, partial assignments to the MIP's variables, guiding towards the optimal solution. 
\tool currently supports the $L_\infty$ perturbation and six semantic feature perturbations (\Cref{fig::intro}(a)). 

We evaluate \tool on several fully-connected and convolutional image classifiers, trained for MNIST, Fashion-MNIST and CIFAR-10. 
We compare to existing approaches, including \emph{Marabou}~\cite{g_ref_31} and sampling-based approaches: \emph{Dataset Sampling}, relying on a dataset, and \emph{Random Sampling}, sampling from the input domain. We further show the importance of integrating the dependencies and hyper-adversarial attacks by comparing them to a MIP-only variant of \tool.
Our results show that (1)~the average gap between the lower and upper bound of \tool is 1.9, while Marabou's gap is 154.7, (2)~the lower bound of \tool is greater (i.e., tighter) by 6.5x than Marabou, by 9.2x than Dataset Sampling and by 25.6x than Random Sampling, (3)~the upper bound of \tool is smaller (i.e., tighter) 
 by 13.1x than Marabou, and (4)~\tool is 130.6x faster than Marabou.
 Our results further indicate that \tool is 78.6x faster than the MIP-only variant. 
 \Cref{fig::intro}(b) shows our results for an MNIST convolutional network, classifying images to digits, and an occlusion perturbation, blackening a 3x3 square at the center of the image. 
 The goal of this experiment is to compute the minimal globally robust bound for which 
 \emph{any input} that the network classifies as the digit 2 with a confidence greater or equal to this bound cannot be perturbed by this occlusion such that the network classifies it as the digit 3. 
 The plot shows that, given a three hour timeout, \tool returns the optimal minimal bound (the lower and upper bound are equal), while the gap of the MIP-only variant is 10.1 and the gap of Marabou is 86.6. The sampling baselines, providing only a lower bound, return a bound 
 lower than the minimal bound by 3x. 
 \Cref{fig::intro}(b) highlights in a light blue background the range of confidences for which every approach proves global robustness, and in a light red background the range of confidences for which every approach guarantees there is no global robustness. \tool perfectly partitions this range, while the other approaches leave a range of confidences without any guarantee.
  %
We further demonstrate how to leverage  \tool's bounds to infer global robustness and vulnerability attributes of the network. 

\section{Preliminaries}
\label{sec:preliminary}
In this section, we provide background on network classifiers. 
We focus on image classifiers.
An image is a $d_1\times d_2$ matrix, where $d_1$ is the height and $d_2$ is the width. The entries are pixels consisting of $l$ channels, each over $[0,1]$. If $l=1$, the image is in grayscale, if $l=3$, it is colored (RGB). 
Given an image domain $[0,1]^{l\times d_1 \times d_2}$ and a set of classes $C=\{1,\dots,c\}$, a classifier maps images to a score vector over the possible classes $D:[0,1]^{l\times d_1\times d_2}\to \mathbb{R}^c$.
We assume a non-trivial classifier, that is $c>1$ and for any class $c'\in C$ there is an input classified as $c'$.
A neural network classifier consists of an input layer followed by $L$ layers. 
For simplicity's sake, our definitions treat all layers as vectors, except where the definition requires a matrix. The input layer $z_0$ takes as input an image 
$x$ and passes it to the next layer (i.e., $z_{0,k}=x_k$).
The last layer outputs a vector, denoted $D(x)$, consisting of a score for each class in $C$. The classification of the network for input $x$ is the class with the highest score, $c'= \text{argmax}(D(x))$.
 The layers are functions, denoted $h_1, h_2, \dots, h_{L}$, each taking as input the output of the preceding layer. The network's function is the composition of the layers: $D(x)=h_{L}(h_{L-1}(\cdots(h_1(x))))$.
A layer $m$ consists of $k_m$ neurons, denoted $z_{m,1},\ldots,z_{m,k_m}$. 
 The function of layer $m$ is defined by its neurons, i.e., it outputs the vector $(z_{m,1},\ldots,z_{m,k_m})^T$.
  There are several types of layers. 
  We focus on fully-connected and convolutional layers, but our approach is easily extensible to other layers such as max pooling layers and residual layers~\cite{HeZRS16}. 
  
  In a fully-connected layer, a neuron $z_{m,k}$ 
   gets as input the outputs of all neurons in the preceding layer. It
  has a weight for each input ${w}_{m,k,k'}$ and a single bias $b_{m,k}$. Its function is computed by first computing the sum of the bias and the multiplication of every input by its respective weight:
 $\hat{z}_{m,k}=b_{m,k}+\sum_{k'=1}^{k_{m-1}}{w}_{m,k,k'}\cdot{z}_{m-1,k'}$. 
 This output is passed to an activation function $\sigma$ to produce the output $z_{m,k}=\sigma(\hat{z}_{m,k})$.
  Activation functions are typically non-linear functions. We focus on ReLU, defined as $\text{ReLU}(\hat{z})=\max(0,\hat{z})$. A convolutional layer is similar to a fully-connected layer except that a neuron does not get as input all neurons from the preceding layer, but only a small subset of them. Additionally, neurons at the same layer share their weights and bias. The set of shared weights and bias are collectively called \emph{a kernel}. Formally, a convolutional layer views its input as a matrix $l\times d^h_{m-1}\times d^w_{m-1}$, where $l$ is the number of input channels, $d^{h}_{m-1}$ is the height and $d^{w}_{m-1}$ is the width.
  A kernel $K$ consists of a weight matrix $W_m$ of dimension $l \times t_m\times t_m$ and a bias $b_m$.
  A kernel is convolved with the input $\hat{z}_{m,l,i,j} = b_m +\sum_{v=1}^{l}\sum^{t_m}_{p=1}\sum^{t_m}_{q=1} W_m[v,p,q] \cdot z_{m-1,v,p,q}$, after which an activation function (e.g., ReLU) is executed component-wise.

\section{Minimal Globally Robust Bounds}\label{sec:problem_definition}

In this section, we present our global robustness property designated for network classifiers. 
 We begin with a few definitions and then define our property and the problem we address.

\paragraph{Perturbations}
A perturbation is a function $f_P(x,\mathcal{E})$, defining how an input $x$ is perturbed given an amount of perturbation specified by a vector $\mathcal{E}=(\epsilon_1,\ldots,\epsilon_k)$. The perturbation values $\epsilon_i$ are either real numbers or integers. 
For example, brightness is defined by a brightness level $\epsilon\in [-1,1]$ and the function is $f_B(x,\epsilon)=x'$ such that $x'_{v,p,q}=x_{v,p,q}+\epsilon$, for $v\in[l]$, $p\in[d_1]$, $q\in[d_2]$.
Translation is defined by translation coordinates $i\in [-d_1,d_1]$ and $j\in[-d_2,d_2]$ and the function is 
$f_T(x,i,j)=x'$ such that $x'_{v,p,q}=(p-i \in [d_1] \wedge q-j\in [d_2]?\ x_{v,p-i, q-j}\ :\ 0)$.
The $L_\infty$ perturbation has a perturbation limit $\epsilon\in[0,1]$ and is defined by a series of perturbations $\epsilon_{v,p,q}\in [-\epsilon,\epsilon]$, for $v\in[l]$, $p\in[d_1]$, $q\in[d_2]$. The function is $f_{L_{\infty}}(x,\epsilon_{1,1,1},\ldots,\epsilon_{l,d_1,d_2})=x'$ such that $x'_{v,p,q}=x_{v,p,q}+\epsilon_{v,p,q}$.
We note that if a perturbed entry is not in its valid range (i.e., below $0$ or above $1$), it is clipped: $\min(\max(f_P(x,\mathcal{E})_{v,p,q},0),1)$. We omit clipping from the formal definitions to avoid a cluttered notation.
A perturbation is confined by a \emph{range} $I_\mathcal{E}$, a series of intervals bounding each entry of $\mathcal{E}$.
For example, a range for brightness is $[0.1,0.3]\subseteq[-1,1]$, bounding the values of the brightness level $\epsilon$, and a 
range for translation is $([1,3],[2,10])\subseteq([-d_1,d_1], [-d_2,d_2])$ (for $d_1\geq 3,d_2\geq10$), bounding the coordinates $i$ and $j$.
An amount of perturbation $\mathcal{E}'$ is in a given range $I_\mathcal{E}$, denoted $\mathcal{E}'\in I_\mathcal{E}$, if its entries are in their intervals: $\forall k.\ \mathcal{E}'_k\in (I_\mathcal{E})_k$. 
For example,
$(2,4)\in ([1,3],[2,10])$.

\paragraph{Local robustness} 
 A common approach 
 to estimate the robustness of a classifier focuses on a finite set of inputs and checks \emph{local robustness} for every input, one-by-one. Local robustness is checked at a given \emph{neighborhood}, defined with respect to an input and a perturbation. Formally, given an input 
$x$, a perturbation $P$ and its range $I_\mathcal{E}$, a neighborhood $\mathcal{N}_{P,I_\mathcal{E}}(x)\subseteq [0,1]^{l\times d_1\times d_2}$ is the set of all inputs obtained by this perturbation: $\mathcal{N}_{P,I_\mathcal{E}}(x)=\{f_P(x,\mathcal{E}')\mid \mathcal{E}' \in I_\mathcal{E}\}$. 
Given a classifier $D$, an input $x$ and a neighborhood containing $x$, $\mathcal{N}_{P,I_\mathcal{E}}(x)\subseteq [0,1]^{l\times d_1\times d_2}$, the classifier $D$ is locally robust at $\mathcal{N}_{P,I_\mathcal{E}}(x)$ if it classifies all inputs the same: $\forall x'\in \mathcal{N}_{P,I_\mathcal{E}}(x).\ \text{argmax}(D(x'))=\text{argmax}(D(x))$. Equivalently, $D$ is locally robust at $x$ under $P$ and $I_\mathcal{E}$ if $\forall \mathcal{E}'\in I_{\mathcal{E}}.\   \text{argmax}(D(f_P(x,\mathcal{E}')))=\text{argmax}(D(x))$.
 \paragraph{Towards a global robustness property}
  Given a perturbation $P$ and its range $I_\mathcal{E}$, an immediate extension of the local robustness definition to global robustness is:
  $$\forall x\forall \mathcal{E}'\in I_{\mathcal{E}}.\   \text{argmax}(D(f_P(x,\mathcal{E}')))=\text{argmax}(D(x))$$ However, no (non-trivial) network classifier and non-trivial perturbation satisfy this property because some inputs $x$ are close to the classifier's decision boundaries~\cite{g_ref_49,g_ref_50} and violate this property. We thus consider only inputs whose network's confidence is high enough and thus are far enough from the decision boundaries. 
  Formally, the network's confidence in classifying an input $x$ as a class $c'\in \{1,\ldots,c\}$ is the difference between the score the network assigns to $c'$ and the maximal score it assigns to any other class. If this confidence is positive, the network classifies $x$ as $c'$, and the higher the (positive) number, the more certain the network is in its classification of $x$ as $c'$. 
\begin{definition}[Class Confidence]
Given a classifier $D$, an input $x$ and a class $c'\in C$, the class confidence of $D$ in $c'$ is $\mathcal{C}(x,c',D)\triangleq D(x)_{c'} - \text{max}_{c''\neq c'}(D(x)_{c''})$.
\end{definition}



\paragraph{$\delta$-Global Robustness}
We define global robustness with respect to a confidence level $\delta$.
The definition restricts the above definition to inputs whose class confidence is at least $\delta$.

\begin{definition}[$\delta$-Globally Robust Classifier]
Given a classifier $D$, a class $c'\in C$, a perturbation ${P}$, a range $I_\mathcal{E}$ and a class confidence $\delta>0$, the classifier $D$ is $\delta$-globally robust for $c'$ under $(P,I_\mathcal{E})$ if:
\begin{align*}
\forall x\forall \mathcal{E}' \in I_{\mathcal{E}}.\ &\mathcal{C}(x,c',D)\geq \delta \Rightarrow  \text{argmax}(D(f_P(x,\mathcal{E}')))=\text{argmax}(D(x))
\end{align*}
\end{definition}
\sloppy
The postcondition $\text{argmax}(D(f_P(x,\mathcal{E}')))=\text{argmax}(D(x))$ is equivalent to $\mathcal{C}(f_P(x,\mathcal{E}'),c',D)>0$, whenever the precondition $\mathcal{C}(x,c',D)\geq \delta$ holds. We thus use these constraints interchangeably.
When it is clear from the context, we write $D$ is $\delta$-globally robust and omit $c'$ and $(P,I_\mathcal{E})$. 
The parameter $\delta$ in the precondition provides the necessary flexibility. The higher the $\delta$, 
 the fewer inputs that are considered and the more likely that the classifier is globally robust with respect to that $\delta$. 
To understand the global robustness level of a classifier, one has to compute the minimal $\delta$  for which the classifier is globally robust. This is the problem we address, which we next define. 


\begin{definition}[Minimal Globally Robust Bound]\label{def:probdef} Given a classifier $D$, a class $c'$, a perturbation ${P}$ and a range $I_\mathcal{E}$,
our goal is to compute a class confidence $\delta_\text{MIN}\in \mathbb{R}^+$ satisfying:
  (1)~$D$ is $\delta_\text{MIN}$-globally robust for $c'$ under $(P,I_\mathcal{E})$, and
  (2)~for every $\delta<\delta_\text{MIN}$, $D$ is not $\delta$-globally robust for $c'$ under $(P,I_\mathcal{E})$.
\end{definition}

This problem is highly challenging for two reasons.
First, determining for a given $\delta$ whether a classifier $D$ is $\delta$-globally robust for $c'$ under $(P,I_\mathcal{E})$ (for non-trivial $\delta$, $D$, $P$ and $I_\mathcal{E}$) requires to determine whether $D$ classifies a very large space of inputs as $c'$. 
Second, to determine the minimal confidence $\delta_\text{MIN}$ for which $D$ is $\delta_\text{MIN}$-globally robust for $c'$ under $(P,I_\mathcal{E})$ requires to determine for a large set of $\delta$ values whether $D$ is $\delta$-globally robust for $c'$ under $(P,I_\mathcal{E})$.

\paragraph{Targeted global robustness} The above definitions aim to show that the network does not change its classification under a given perturbation. In some scenarios, network designers may permit some class changes. For example, they may allow the network to misclassify a perturbed dog as a cat, but not as a truck. 
Thus, when determining global robustness, it is sometimes more useful to guarantee that the network does not change its classification to a target class, rather than to any class but the correct class. We next adapt our definitions to targeted global robustness.  

\begin{definition}[$\delta$-Targeted Globally Robust Classifier]\label{def:targeted}
Given a classifier $D$, a class $c'\in C$, a target class $c_t\in C$ such that $c_t\neq c'$, a perturbation ${P}$, a range $I_\mathcal{E}$ and a class confidence $\delta>0$, the classifier $D$ is $\delta$-globally robust \emph{against} $c_t$ under $(P,I_\mathcal{E})$ if:
\begin{align*}
\forall x\forall \mathcal{E}' \in I_{\mathcal{E}}.\ &\mathcal{C}(x,c',D)\geq \delta \Rightarrow   \text{argmax}(D(f_P(x,\mathcal{E}')))\neq c_t
\end{align*}
\end{definition}
The problem definition of computing the minimal targeted globally robust bound for a target class $c_t$ is the same as \Cref{def:probdef} but with respect to \Cref{def:targeted}. 
In the next section, we continue with the untargeted definition (\Cref{def:probdef}), but all definitions and algorithms easily adapt to the targeted definition.  
\section{Overview of Key Ideas}

\label{sec:framework}
\begin{figure*}[t]
    \centering
  \includegraphics[width=1\linewidth, trim=0 335 0 0, clip,page=7]{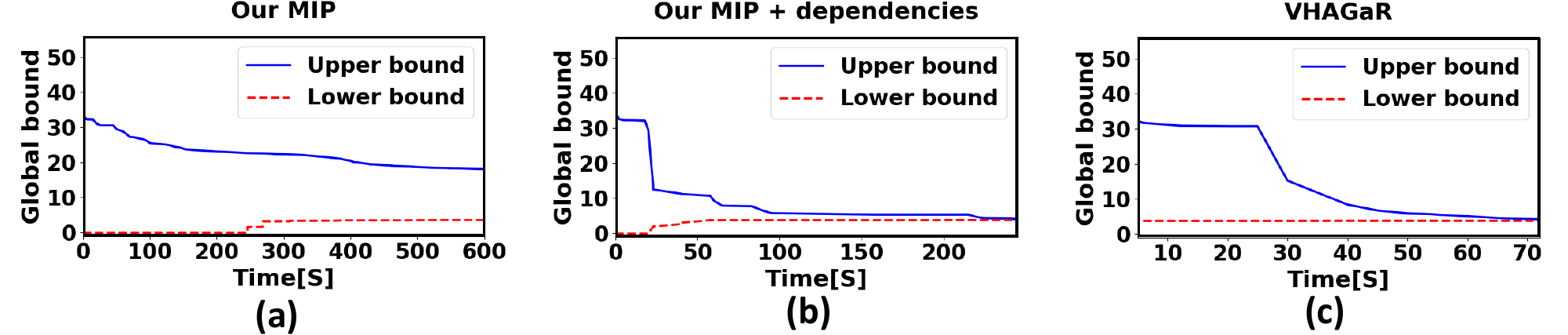}
    \caption{The bounds as a function of the execution time,
     when providing a MIP solver (a)~our encoding, (b)~additionally our dependencies, and (c)~additionally our suboptimal lower bound and optimization hints.}
     
    \label{fig::example_dep_and_feasible1}
\end{figure*}

In this section, we present our main ideas to compute the minimal globally robust bound. We begin by
introducing the straightforward optimization formalism for our problem and explaining why it is not amenable to standard MIP solvers. Then, we introduce the key components of \tool. 
First, we rephrase the optimization problem into a form supported by MIP solvers, which also enables \tool an anytime optimization (\Cref{sec:key_idea_new_form}). 
The anytime computation splits the problem into computing an upper bound and a lower bound for the minimal globally robust bound. 
Second, we identify dependencies that reduce the time complexity of solving the optimization problem (\Cref{sec:key_idea_dependencies}). 
Third, we compute a suboptimal lower bound and optimization hints, using a \emph{hyper-adversarial attack}, which are provided to the MIP solver to expedite the tightening of the bounds (\Cref{sec:key_idea_feasible_solutions}). 
\Cref{fig::example_dep_and_feasible1} demonstrates the importance of our steps. 
This figure shows the upper and lower bounds obtained by the anytime optimization as a function of the time, for a convolutional network composed of 540 neurons trained for the MNIST dataset and an occlusion perturbation. 
\Cref{fig::example_dep_and_feasible1}(a) shows the bounds obtained by our encoding (\Cref{sec:key_idea_new_form}). While the bounds are tightened given more time, their gap is quite large after ten minutes. 
\Cref{fig::example_dep_and_feasible1}(b) shows the improvement obtained by encoding the dependencies (\Cref{sec:key_idea_dependencies}): the bounds are tightened more significantly and more quickly.
\Cref{fig::example_dep_and_feasible1}(c) shows the improvement obtained by additionally providing the suboptimal lower bound and the optimization hints (\Cref{sec:key_idea_feasible_solutions}): the lower and upper bounds are tightened more quickly and converge to the minimal globally robust bound within 72 seconds.

\subsection{Minimal Globally Robust Bound via Optimization}
\label{sec:key_idea_new_form}
In this section, we phrase the problem of minimal globally robust bound as constrained optimization. 
Phrasing a problem as constrained optimization and providing it to a suitable solver has been shown to be very efficient in solving complex search problems and in particular robustness-related tasks, such as computing adversarial examples~\cite{g_ref_9,g_ref_4,g_ref_5,g_ref_6}, proving local 
 robustness~\cite{g_ref_33,g_ref_39,g_ref_28,g_ref_23} and computing maximally robust neighborhoods~\cite{g_ref_53,g_ref_56,g_ref_55,g_ref_54}. 
However, the straightforward formulation of our task, building on the aforementioned tasks, is not amenable to existing MIP solvers. We thus present an equivalent formulation, amenable to these solvers.
We begin with the straightforward formulation and then show the equivalent formulation. 
The straightforward formulation follows directly from \Cref{def:probdef}:
\begin{problem}[Minimal Globally Robust Bound]
Given a classifier $D$, a class $c'$, a perturbation $P$ and a range $I_{\mathcal{E}}$, the constrained optimization computing the minimal globally robust bound is:
\begin{equation}\label{problem1}
  \begin{gathered}
     \min \delta \hspace{0.25cm}
     \text{subject to}\hspace{0.25cm}
     \forall x\forall \mathcal{E}'\in I_{\mathcal{E}}.\ \mathcal{C}(x,c',D)\geq\delta \Rightarrow
\mathcal{C}(f_P(x,\mathcal{E}'),c',D)>0
    \end{gathered}
\end{equation}

\end{problem}


This constraint requires that for any input classified as $c’$ with confidence greater or equal to $\delta$, its perturbations are classified as $c’$. 
The optimal solution $\delta^*_{(1)}$ ranges over $[\Delta,\delta^m_{c'}+\Delta]$, where $\delta^m_{c'}$ is the maximal confidence that the classifier $D$ assigns to $c'$ for any input (i.e., $\delta^m_{c'}=\max_x\;\mathcal{C}(x,c',D)$) and $\Delta$ is a very small number that denotes the computer's precision level for representing the real numbers. 
For example, if the perturbation is the identity function (i.e., $f_P(x,\mathcal{E}')=x$), then $\delta^*_{(1)}=\Delta$.
More generally, if the perturbation does not lead $D$ to change the classification for any input, then $\delta^*_{(1)}=\Delta$.
A different example is a perturbation that can perturb every pixel within its range $[0,1]$.
In this case, $\delta^*_{(1)}=\delta^m_{c'}+\Delta$. This follows since, under this perturbation,
the right-hand side of the constraint pertains to every input (i.e., for every $x,x'\in[0,1]^{l\times d_1\times d_2}$ there is $\mathcal{E}'$ such that $f_P(x,\mathcal{E'})=x'$). Since we assume $D$ is a non-trivial classifier that has at least two classes ($c>1$), every input $x$ violates the postcondition because there are inputs $x'$ not classified as $c'$. Namely, $\delta^m_{c'}$ is the maximal value for which the postcondition does not hold.  
Since for $\delta^m_{c'}+\Delta$ the precondition is vacuously true, $\delta^*_{(1)}=\delta^m_{c'}+\Delta$ is the minimal value satisfying the constraint of Problem~\ref{problem1}.
For other (more interesting) perturbations, $\delta^*_{(1)}\in (\Delta,\delta^m_{c'}+\Delta)$ (note that $\delta^m_{c'}$ depends on $D$).

Problem~\ref{problem1} cannot be submitted to existing MIP solvers, because its constraint has a for-all operator defined over a continuous range. Instead, we consider the dual problem: computing the \emph{maximal globally non-robust bound}. 
This problem looks for the {maximal} ${\delta}$ \emph{violating} the constraint of Problem~\ref{problem1}. Namely, for every $\delta'>{\delta}$ the network is $\delta'$-globally robust, and for every $\delta'\leq{\delta}$ there is an input $x$ and a perturbation amount $\mathcal{E}'$ for which the network is not robust. Formally:

\begin{problem}[Maximal Globally Non-Robust Bound]
\begin{equation}\label{problem2}
  \begin{gathered}
     \max \delta \hspace{0.25cm}
     \text{subject to}\hspace{0.25cm}
     \exists x\exists \mathcal{E}'\in I_{\mathcal{E}}.\ \mathcal{C}(x,c',D)\geq \delta \wedge \mathcal{C}(f_P(x,\mathcal{E}'),c',D)\leq 0
    \end{gathered}
\end{equation}
\end{problem}

In the special case where the perturbation does not lead $D$ to change its classification for any input, i.e., $\delta^*_{(1)}=\Delta$, Problem~\ref{problem2} is infeasible. For completeness of the definition, in this case we define $\delta^*_{(2)}\triangleq 0$. 
Thus, by definition, the optimal solution $\delta^*_{(2)}$ ranges over $[0,\delta^m_{c'}]$. 
Note that Problem~\ref{problem1} and Problem~\ref{problem2} have very close optimal solutions: their difference is the precision level, namely $\delta^*_{(1)}-\delta^*_{(2)}=\Delta$. This is because $\delta^*_{(1)}$ is the minimal confidence for which there is no adversarial example and $\delta^*_{(2)}$ is the maximal confidence for which there is an adversarial example. Formally:  
%
\begin{restatable}[]{lemma}{fta}
\label{lem1}
Let $\delta^*_{(1)}$ be the optimal value of Problem~\ref{problem1} and $\delta^*_{(2)}$ be the optimal value of Problem~\ref{problem2} 
(where $\delta^*_{(2)}\triangleq 0$ if Problem~\ref{problem2} is infeasible).
Then $\delta^*_{(1)} -\delta^*_{(2)} =\Delta$, where $\Delta$ is the precision level.

\end{restatable}

\begin{proof} 
If Problem~\ref{problem2} is infeasible, then $\delta^*_{(2)}\triangleq 0$. In this case, all perturbed examples are classified as $c'$ and by definition of Problem~\ref{problem1}, $\delta^*_{(1)}=\Delta$. 
  If Problem~\ref{problem2} is feasible, then since it is bounded (by $\delta^m_{c'}$), it has an optimal solution
  $\delta^*_{(2)}$.
  If we add to $\delta^*_{(2)}$ the smallest possible value $\Delta$, we obtain  $\delta^*_{(2)}+ \Delta$, which violates the constraint of Problem~\ref{problem2}. Since Problem~\ref{problem1}'s constraint is the negation of Problem~\ref{problem2}'s constraint, $\delta^*_{(2)}+ \Delta$ satisfies the constraint of Problem~\ref{problem1} and is thus a feasible solution. It is also the minimal value satisfying this constraint: subtracting the smallest value possible $ \Delta $ results in $\delta^*_{(2)}$, which does not satisfy Problem~\ref{problem1}'s constraint. Thus, $\delta^*_{(1)}=\delta^*_{(2)}+ \Delta$. 
\end{proof}

In the following, we sometimes abuse terminology and say that the optimal solution of Problem~\ref{problem2}, $\delta^*_{(2)}$, is an optimal solution to 
Problem~\ref{problem1}, because given $\delta^*_{(2)}$, we can obtain $\delta^*_{(1)}$ by adding the precision level $\Delta$.  
Besides having close optimal solutions, the problems have the same time complexity, which is very high (as we shortly explain). The main difference between these problems is that the second formulation is supported by existing MIP solvers. A common and immediate practical approach to address highly complex problems is by anytime optimization. We next discuss at high-level the time complexity and how we leverage the anytime optimization. 

\paragraph{Problem complexity} \tool solves Problem~\ref{problem2} by encoding the problem as a MIP problem (the exact encoding is provided in~\Cref{sec:the_mip_encoding}) and then submitting it to a MIP solver. The encoding includes the network's computation twice: once as part of the encoding of the class confidence of $x$ and once as part of the encoding of the class confidence of $f_P(x,\mathcal{E}')$. Each network's computation is expressed by the encoding proposed by~\citet{g_ref_33} for local robustness analysis.
This encoding assigns to each non-linear computation of the network (e.g., ReLU) a boolean variable, whose domain is $\{0,1\}$.
Generally, the time complexity of a MIP problem over boolean and real variables is exponential in the number of boolean variables.
 To make the analysis more efficient, \citet{g_ref_33} identify boolean variables that can be removed (e.g., ReLUs whose inputs are non-positive or non-negative), thereby reducing the complexity.
 Despite the exponential complexity, MIP encoding is popular for verifying local robustness~\cite{g_ref_33,g_ref_39,g_ref_28,g_ref_23}.
 However, for global robustness, where the network's computation is encoded twice, the exponent is twice as large. This leads to even larger complexity, posing a significant challenge for the MIP solver. 
 To enable the MIP solver to reach a solution for our optimization problem, we propose several steps, the first one is a built-in feature of existing solvers, and we next describe it. 

\paragraph{Anytime optimization}
\label{sec:key_idea_anytime}
 One immediate mitigation for the high time complexity is to employ anytime optimization. 
We next describe how Gurobi, a state-of-the-art optimizer, supports anytime optimization. 
To find the optimal solution to Problem~\ref{problem2}, $\delta^*_{(2)}$, Gurobi defines two bounds: an upper bound $\delta_U\geq \delta^*_{(2)}$ and a lower bound $\delta_L\leq \delta^*_{(2)}$. The upper bound is initialized by a large value, e.g., $\delta_U=\text{MAX\_FLOAT}$, that satisfies the constraint of our original optimization problem (Problem~\ref{problem1}).
It then
 iteratively decreases $\delta_U$ until reaching a value violating this constraint. For the lower bound, Gurobi looks for feasible solutions for our optimization problem (Problem~\ref{problem2}). That is, it looks for inputs classified as $c'$ and corresponding perturbations $\mathcal{E}'$ that cause the network to misclassify. 
 These parallel updates continue until $\delta_U=\delta_L$. This modus operandi provides a natural anytime optimization: 
 at any time the user can terminate the optimization and obtain that $\delta^*_{(2)}\in [\delta_L,\delta_U]$. 
 That is, an anytime optimization provides a practical approach to bound the network's maximal globally non-robust bound (and hence bound the minimal globally robust bound $\delta^*_{(1)}\in [\delta_L+\Delta,\delta_U+\Delta]$).
 Combined with our other steps to scale the optimization, \tool
 often reaches relatively tight bounds (i.e., the difference $\delta_U-\delta_L$ is small), as we show in~\Cref{sec:eval}. 


\subsection{Reducing Complexity via Encoding of Dependencies}
\label{sec:key_idea_dependencies}
In this section, we explain our main idea for reducing the time complexity of our optimization problem: encoding dependencies stemming from the perturbation or the network's computation. This step expedites the time to tighten the bounds of the optimal solution.
Recall that the optimization problem has a constraint over the class confidence of an input $\mathcal{C}(x,c',D)$ and the class confidence of its perturbed example $\mathcal{C}(f_P(x,\mathcal{E}'),c',D)$. 
The perturbation function $f_P$ imposes relations between the input $x$ and the perturbed input $f_P(x,\mathcal{E}')$. For some perturbations, these relations can be encoded as linear or quadratic constraints. Additionally, since our encoding captures the network's computation for an input and its perturbed example, the outputs of respective neurons may also be linearly dependent via equality or inequality constraints. Adding these dependencies to our MIP encoding can significantly reduce the problem's complexity, as we next explain.


As described, the time complexity is governed by the number of boolean variables in the MIP encoding. \tool relies on the encoding by \citet{g_ref_33} in which every neuron has a unique real-valued variable and every non-input neuron (which executes ReLU) has a unique boolean variable. Since \tool encodes the network twice, once for the input and once for the perturbed example, every neuron has two real-valued variables and two boolean variables. The real-valued variables of neuron $k$ in layer $m$ are $z_{m,k}$ for the input and $z^p_{m,k}$ for the perturbed example, and they capture the computation defined in~\Cref{sec:preliminary}. The boolean variables are $a_{m,k}$ for the input and $a^p_{m,k}$ for the perturbed example. 
If $a_{m,k} = 1$ (respectively $a_{m,k}^p=1$), then this neuron's ReLU is active, $z_{m,k}\geq 0$ (respectively $z_{m,k}^p\geq 0$); otherwise, it is inactive, $z_{m,k}\leq 0$ (respectively $z_{m,k}^p\leq 0$). 
~Several works propose ways to reduce the number of boolean variables~\citep{g_ref_40,g_ref_39,g_ref_1,g_ref_21}. 
Two common approaches are: (1)~computing the concrete lower and upper bounds of ReLU neurons, by submitting additional optimization problems to the MIP solver, to identify ReLU computations that are in a stable state (i.e., their inputs are non-positive or non-negative) and (2)~over-approximating the non-linear ReLU computation using linear constraints. 
We show how to compute equality and inequality constraints over respective variables ($z_{m,k},z_{m,k}^p$ and $a_{m,k},a_{m,k}^p$), stemming from the network's computation, via additional optimization problems (described in~\Cref{sec:semantic_perturbations}). 
Additionally, we identify dependencies that stem from the perturbation and do not involve additional optimization, as we next explain.

\begin{figure*}[t]
    \centering
  \includegraphics[width=1\linewidth, trim=0 220 30 0, clip,page=6]{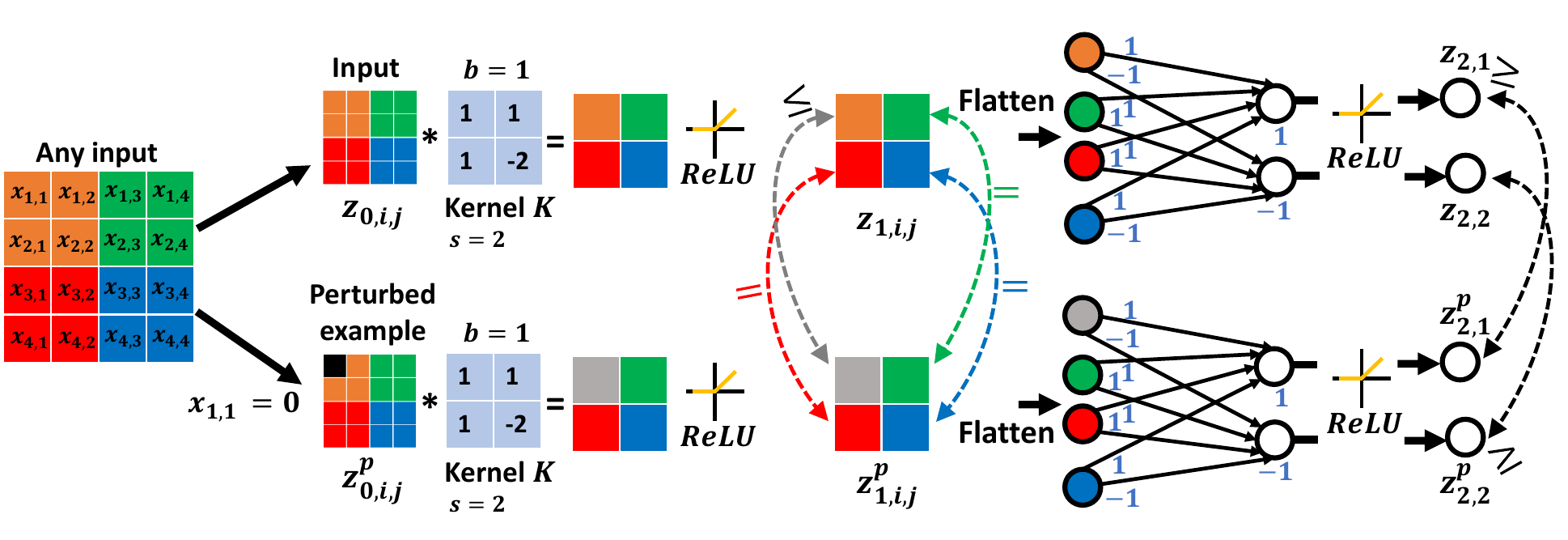}
    \caption{An example of the perturbation dependencies of an occlusion perturbation blackening the pixel $(1,1)$.}
    \label{fig::conv_and_fully_global}
\end{figure*}

Many perturbations impose relations expressible as linear or quadratic constraints over the input and its perturbed example. 
These constraints allow pruning the search space. In particular, sometimes the dependencies pose an equality constraint over two boolean variables, effectively reducing the number of boolean variables, and thereby reducing the problem's complexity. We define the perturbation dependencies formally in \Cref{sec:In_dep}, after describing the MIP encoding. We next demonstrate at a high-level the perturbation dependency of the occlusion perturbation.
\paragraph{Example}
 Consider
\Cref{fig::conv_and_fully_global}, showing an example of a small network comprising 16 inputs $z_{0,i,j}\in[0,1]$, two outputs $o_1,o_2$, and two hidden layers. The first hidden layer is a convolutional layer composed from four ReLU neurons. The second hidden layer is a fully-connected layer composed from two ReLU neurons; its weights are depicted by the edges and its biases by the neurons. Consider the occlusion perturbation, blackening the pixel (1,1). 
Blackening it means setting its value to zero (i.e., the minimal pixel value). 
The figure shows, at the top, the input network, receiving the input, and, at the bottom, the perturbation network,  receiving the perturbed example. 
By the encoding of \citet{g_ref_33}, 
each neuron of the input network has a real-valued variable, denoted $z_{m,i,j}$ for $m\in\{0,1\}$ (for the input layer and the convolutional layer), or $z_{m,k}$ for $m=2$ (for the fully-connected layer), where $i$, $j$, and $k$ are the indices within the layers. Each non-input neuron also has a boolean variable, denoted $a_{1,i,j}$ or $a_{2,k}$.
Similarly, each neuron of the perturbation network has a real-valued variable, denoted $z^p_{m,i,j}$ for $m\in\{0,1\}$ and $z^p_{2,k}$, and each non-input neuron has a boolean variable, denoted $a^p_{1,i,j}$ or $a^p_{2,k}$.
Because the occlusion perturbation changes a single input pixel, all input neurons besides $(1,1)$ are the same: $z_{0,i,j}=z^p_{0,i,j}$, for $(i,j)\in [4]\times[4]\setminus \{(1,1)\}$.
Consequently, the computations of the subsequent layers are also related.
In the convolutional layer, the neurons at indices $(1,2),(2,1),(2,2)$ accept the same values in the input and the perturbed networks, and the neuron at index (1,1) accepts smaller or equal values in the perturbed network compared to the input network (recall $z^p_{0,1,1}=0\leq z_{0,1,1}$). 
Thus, \tool adds the dependencies: $z_{1,i,j}=z^p_{1,i,j}$ and $a_{1,i,j}=a^p_{1,i,j}$ for $(i,j)\in\{(1,2),(2,1),(2,2)\}$ as well as $z_{1,1,1}\geq z^p_{1,1,1}$ and $a_{1,1,1} \geq a^p_{1,1,1}$. 
 Similarly, in the fully-connected layer, \tool adds the dependencies $z_{2,1}\geq z^p_{2,1}$ and $a_{2,1} \geq a^p_{2,1}$, for the first neuron, and $z_{2,2}\leq z^p_{2,2}$ and $a_{2,2} \leq a^p_{2,2}$, for the second neuron. Adding these constraints to the MIP encoding prunes the search space. Further, it effectively removes three boolean variables.

\subsection{Computing a Suboptimal Lower Bound by a Hyper-Adversarial Attack}
\label{sec:key_idea_feasible_solutions}

In this section, we present our approach to expedite the tightening of the lower bound by efficiently computing a suboptimal feasible solution to Problem~\ref{problem2} and providing it to the MIP solver. 
Computing feasible solutions amounts to computing adversarial examples. This is because feasible solutions are inputs $x$ that have a perturbation amount $\mathcal{E}'$ that leads the network to change its classification. 
We present a \emph{hyper-adversarial attack}, generalizing existing adversarial attacks to unknown inputs. 
A hyper-adversarial attack leverages numerical optimization, which is more effective for finding adversarial examples, and thereby it expedites the lower bound computation of the general-purpose MIP solver.  
We also use the hyper-adversarial attack to provide the MIP solver \emph{hints} (a partial assignment) for the optimization variables to further guide the MIP solver to the optimal solution. 

\begin{figure}[t]
    \centering
  \includegraphics[width=1\linewidth, trim=0 0 0 0, clip,page=14]{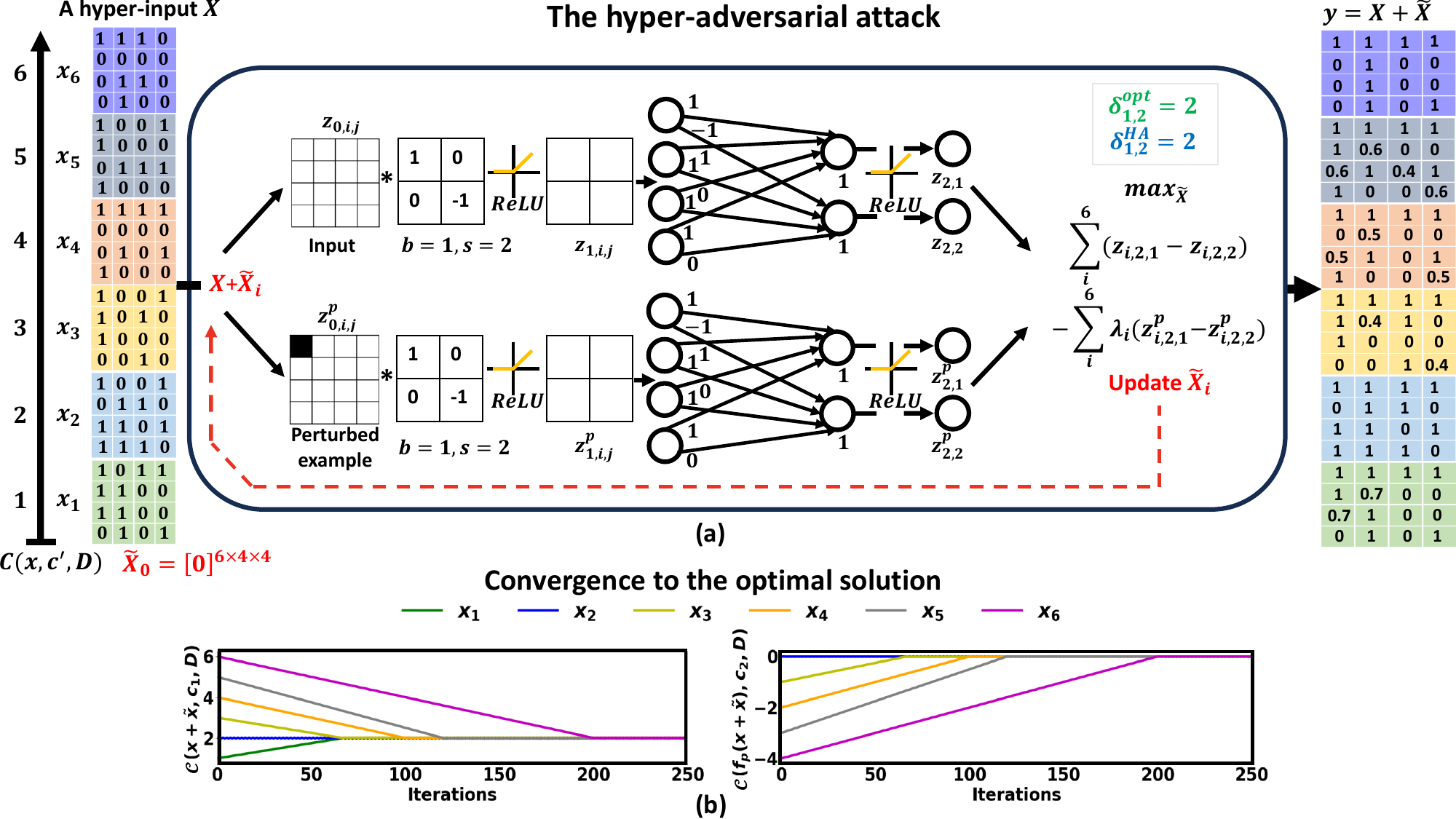}
    \caption{An example of running the hyper-adversarial attack to obtain a suboptimal lower bound on the maximal globally non-robust bound for the occlusion perturbation blackening the pixel (1, 1).}
    \label{fig::conv_and_fully_global_hyper_attacks}
\end{figure}

An adversarial attack gets an input and computes a perturbation that misleads the network. In our global robustness setting, there is no particular input to attack. Instead, the space of feasible solutions (i.e., inputs) is defined by the constraint: $\exists x\exists \mathcal{E}'\in I_{\mathcal{E}}.\ \mathcal{C}(x,c',D)\geq \delta 
\wedge \mathcal{C}(f_P(x,\mathcal{E}'),c',D)\leq 0$.
      A hyper-adversarial attack begins by generating a \emph{hyper-input}. A hyper-input is a set of $M$ inputs $X=\{ x_1,x_2,x_3,...,x_M\}$ classified as $c'$. 
      It consists of inputs with a wide range of class confidences. 
      This property enables the hyper-adversarial attack to quickly converge to a non-trivial suboptimal (or optimal) lower bound. This is because the large variance in the confidences means that some inputs are closer 
      to the optimal lower bound. These inputs provide the optimizer with a good starting point to begin searching for the optimal value \emph{without knowing the optimal value}. We shortly exemplify this property (\Cref{fig::conv_and_fully_global_hyper_attacks}(b)).
       Given a hyper-input $X$, a hyper-adversarial attack simultaneously looks for inputs that have adversarial examples and their class confidence is higher than those of the inputs in $X$. Unlike existing adversarial attacks, a hyper-adversarial attack does not look for a perturbation that causes misclassification but rather for \emph{inputs} that have perturbations that cause misclassification.
       Formally, given a hyper-input $X$, the hyper-adversarial attack is defined over input variables $\tilde{X}=(\tilde{x}_1,\ldots,\tilde{x}_M)^T$ and perturbation variables $\tilde{\mathcal{E}}'=(\tilde{\mathcal{E}}'_1,\ldots,\tilde{\mathcal{E}}'_M)^T$, each corresponding to an input in $X$. The goal of the search is to find a vector $\tilde{X}$ that when added to the vector $X$ leads to at least one input with a higher class confidence that has an adversarial example. The corresponding optimization problem is the following multi-objective optimization problem:

\begin{equation}\label{problem3}
  \begin{gathered}
      \max_{\tilde{X}} \mathcal{C}(x_1+\tilde{x}_1,c',D),..., \mathcal{C}(x_M+\tilde{x}_M,c',D)\;\hspace{0.12cm}
     \text{subject to}
     \bigvee_{ i \in [M]}\ \exists \tilde{\mathcal{E}}'_i.\ \mathcal{C}(f_P(x_i+\tilde{x}_i,\tilde{\mathcal{E}}'_i),c',D)\leq0
    \end{gathered}
\end{equation}

Since all inputs in $X$ have a positive class confidence for $c'$, 
 we can replace the $M$ maximization objectives with a single equivalent objective defined as the sum over the $M$ individual objectives. By aggregating the objectives with a sum, we streamline the optimization process. To make this constrained optimization amenable to standard optimization approaches, we strengthen the constraint and replace the disjunction by a conjunction.
To solve the constrained optimization, we employ a standard relaxation (e.g., \cite{g_ref_4,g_ref_5,g_ref_51,g_ref_9}) and transform it into an unconstrained optimization by presenting $M$ optimization values $\lambda = \{\lambda_1,\ldots,\lambda_M\}$ and adding the conjuncts as additional terms to the optimization goal. Overall, \tool optimizes the following loss function:
$$ \max_{\tilde{X},\tilde{\mathcal{E}}'}\sum_{i=1}^M{\mathcal{C}(x_i+\tilde{x}_i,c',D)} - \sum_{i=1}^M{\lambda_i\cdot\mathcal{C}(f_P(x_i+\tilde{x}_i,\tilde{\mathcal{E}}'_i),c',D)}$$
 \tool maximizes this loss by adapting the PGD attack~\cite{g_ref_6} to consider adaptive $\lambda_i$ (explained in~\Cref{sec:sub_opt_sol_and_opt_hints}). 
A solution to this optimization is an assignment to the variables, from which \tool obtains a set of solutions denoted as $y=(y_1,\ldots,y_M)$, where $y_i=x_i+\tilde{x}_i$. 
\tool looks for the input $y^* \in y$ maximizing the class confidence. It then provides its confidence to the MIP solver as a lower bound for our optimization problem (Problem~\ref{problem2}). Additionally, 
it provides the MIP solver \emph{optimization hints}. A hint is an assignment to an optimization variable from which the MIP solver proceeds its search.
\tool computes hints for the boolean variables (i.e., $a_{m,k},a^p_{m,k}$), which play a significant role in the problem's complexity. 
It does so by first running the feasible solutions and their perturbed examples through the network to obtain their assignment to the optimization variables. Then, it defines a hint for every boolean that has the same value for the majority of the feasible solutions. 
We rely on the majority to avoid biasing the optimization to a single feasible solution's values.   
Next, we provide an example of our hyper-adversarial attack. 

\paragraph{Example} 
\Cref{fig::conv_and_fully_global_hyper_attacks}(a) shows a small network, which has 16 inputs $z_{0,i,j}\in[0,1]$, two outputs $o_1,o_2$ defining two classes $C=\{c_1, c_2\}$, and two hidden layers. 
The weights are shown by the edges and the biases by the neurons. 
We consider an occlusion perturbation that blackens the pixel (1,1). For simplicity's sake, we consider a constant perturbation, which allows us to omit the perturbation variables $\tilde{\mathcal{E}}'$.
In this example, the hyper-adversarial attack looks for inputs maximizing the class confidence of $c'=c_1$, whose perturbed examples are misclassified as $c_2$. 
\tool begins from a hyper-input consisting of six inputs $X=(x_1,\ldots,x_6)$ classified as $c_1$ with the confidences $\{1,\dots,6\}$, respectively (\Cref{fig::conv_and_fully_global_hyper_attacks}(a), left). Note that these inputs' confidences are uniformly distributed over the range of confidences $[1,6]$, providing a wide range of confidences. 
\tool then initializes for every input a respective variable $\widetilde{X}=(\tilde{x}_1,\ldots,\tilde{x}_6)$. Initially, $\tilde{x}_i=0$, for every $i$.
\tool then runs the PGD attack over the loss: 
$\max_{\tilde{X}}\sum_{i=1}^6{\mathcal{C}(x_i+\tilde{x}_i,c',D)} - \sum_{i=1}^6{\lambda_i\mathcal{C}(f_P(x_i+\tilde{x}_i,\mathcal{E}'),c',D)}$, where $f_P$ and $\mathcal{E}'$ capture blackening the pixel $(1,1)$.
At a high-level, this attack runs iterations. 
An iteration begins by running each input $x_i+\tilde{x}_i$ through the network to compute $\mathcal{C}(x_i+\tilde{x}_i,c',D)$, and 
each $f_P(x_i+\tilde{x}_i,\mathcal{E}')$, which is identical to $x_i+\tilde{x}_i$ except for the black pixel, through the network to compute $\mathcal{C}(f_P(x_i+\tilde{x}_i,\mathcal{E}'),c',D)$.
   Then, \tool computes the gradient of the loss, updates $\widetilde{X}$ based on the gradient direction, and begins a new iteration, until it converges.
    On the right, the figure shows the inputs at the end of the optimization. In this example, all the respective perturbed examples are classified as $c_2$ and the maximal class confidence of $c_1$ is $2$.
        The value $2$ is submitted to the MIP solver as a lower bound on the maximal globally non-robust bound.
    This lower bound prunes for the MIP solver many exploration paths leading to inputs with class confidence lower than 2. For example, it prunes every exploration path in which $a_{1,i,j}=0$, for every $i$ and $j$. Thereby, it prunes every assignment in the following space:  $\forall i \forall j.\ a_{1,i,j}=0 \land \forall i.\ a_{2,i}\in\{0,1\} \land \forall i \forall j.\  a^p_{1,i,j}\in\{0,1\} \land \forall i.\ a^p_{2,i}\in\{0,1\}$. This pruning is obtained by bound propagation, a common technique in modern solvers. 
In addition to the lower bound, \tool computes optimization hints by looking for boolean variables with the same assignment, for the majority of inputs or their perturbed examples. In this example, the hints are $\forall i\forall j.\ a_{1,i,j}=a^p_{1,i,j}=1$ and $\forall i.\ a_{2,i}=a_{2,i}^p=1$. 
    These hints are identical to the boolean values corresponding to the optimal solution of Problem~\ref{problem2}. 
    By providing the hints to the MIP solver, it converges efficiently to a suboptimal (or optimal) solution.

     \Cref{fig::conv_and_fully_global_hyper_attacks}(b) shows the change in the class confidences $\mathcal{C}(x_i+\tilde{x}_i,c_1,D)$ (left) and 
      $\mathcal{C}(f_P(x_i+\tilde{x}_i,\mathcal{E}'),c_2,D)=-\mathcal{C}(f_P(x_i+\tilde{x}_i,\mathcal{E}'),c_1,D)$~(right) during the optimization, for all six inputs. 
    The left plot shows that initially, since $\tilde{x}_i=0$, the confidences of $x_i+\tilde{x}_i$ are uniformly distributed. As the optimization progresses, the $\tilde{x}_i$-s are updated and eventually all $x_i+\tilde{x}_i$ converge to the 
    optimal maximal globally non-robust bound, which is $2$. 
    Note that inputs that initially have close confidences to the maximal globally non-robust bound converge faster than inputs that initially have a confidence that is farther from the bound. 
    This demonstrates the importance of running the hyper-adversarial attack over inputs with a wide range of confidences: it enables to expedite the search for a feasible solution with a high confidence (we remind that the optimal bound is unknown to the optimizer and it may terminate before converging to all inputs). In particular, it increases the likelihood of reaching a suboptimal lower bound.
    The right plot shows that all perturbed examples are eventually classified as $c_2$.

 \begin{figure*}[t]
    \centering
  \includegraphics[width=1\linewidth, trim=0 80 0 0, clip,page=1]{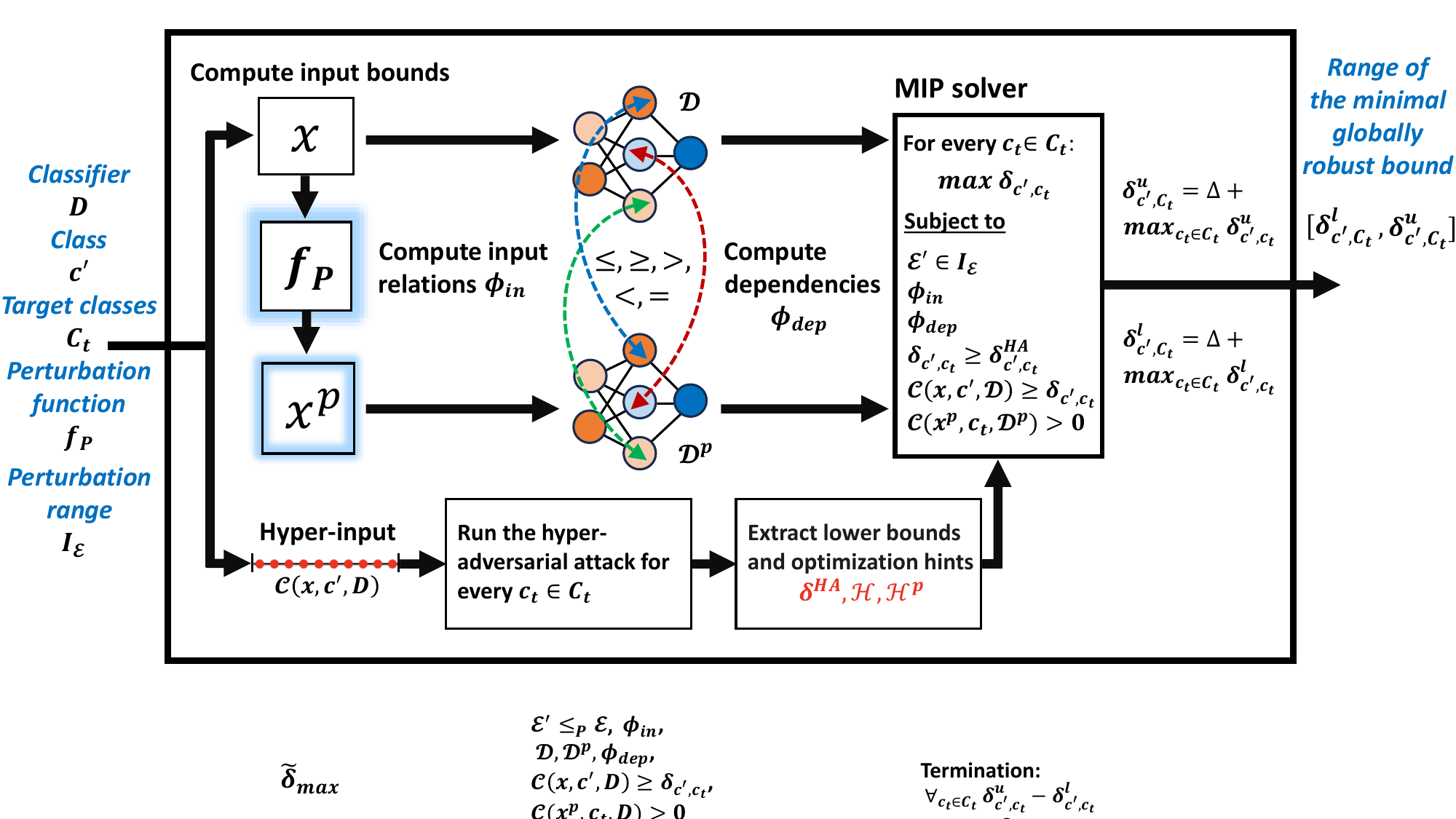}
    \caption{Illustration of \tool. }
    \label{fig::all_system}
\end{figure*} 

\section{VHAGaR: a Verifier of Hazardous Attacks for Global Robustness}
In this section, we present \tool, our anytime system for computing an interval bounding the minimal globally robust bound. 
\Cref{fig::all_system} illustrates its operation.
\tool takes as arguments a classifier $D$, a class $c'$, a set of target classes $C_t\subseteq C\setminus\{c'\}$, a perturbation function $f_P$, and a perturbation range $I_\mathcal{E}$. 
 Note that \tool takes as input a \emph{set of target classes}, thereby it can support the untargeted definition (where $C_t=C\setminus\{c'\}$), the (single) targeted definition (where $C_t=\{c_t\}$ for some $c_t\neq c'$), or any definition in between (where $C_t$ contains several classes).
\tool begins by encoding Problem~\ref{problem2} as a MIP. This encoding consists of two copies of the network's computation, $\mathcal{D}$ for the input $x$ and $\mathcal{D}^p$ for the perturbed example $x^p$. To scale the MIP solver's computation, \tool computes real-valued lower and upper bounds by adapting the approach of~\citet{g_ref_33}.
\tool then adds a constraint $\phi_{in}$ capturing the dependencies stemming from the perturbation and 
 a constraint $\phi_{dep}$ capturing dependencies over the two networks' neurons. 
In parallel to the MIP encoding, 
\tool computes for each class in $C_t$ a suboptimal lower bound $\delta^{HA}$ by running the hyper-adversarial attack. 
From the attack, it also obtains optimization hints and passes them to the MIP solver in two matrices, $\mathcal{H}$ for $\mathcal{D}$ and $\mathcal{H}^p$ for $\mathcal{D}^p$.  
We note that in our implementation, the MIP encoding runs on CPUs, and in parallel the hyper-adversarial attack runs on GPUs. 
After the MIP encoding and the hyper-adversarial attack complete, \tool submits for every target class in $C_t$ a MIP to the MIP solver (in parallel). 
The solver returns the lower and upper bound for each MIP, computed given a timeout.
The maximum lower and upper bounds over these bounds form 
the interval of the maximal globally non-robust bound.
\tool adds the precision level $\Delta$ and returns
this interval, bounding the minimal globally robust bound. 
 We next present the MIP encoding (\Cref{sec:the_mip_encoding}), the dependency computation (\Cref{sec:semantic_perturbations}), and the hyper-adversarial attack (\Cref{sec:sub_opt_sol_and_opt_hints}). Finally, we discuss correctness and running time (\Cref{sec:correctness_and_running_time}).
\label{sec:milp_encoding}
\subsection{The MIP Encoding}
\label{sec:the_mip_encoding}

In this section, we present \tool's MIP encoding which adapts the encoding of \citet{g_ref_33} for local robustness verification. 
To understand \tool's encoding, we first provide background on encoding local robustness as a MIP, for a fully-connected classifier (the extension to a convolutional network is similar). Then, we describe our adaptations for our global robustness setting.

\paragraph{Background} \citet{g_ref_33} propose a sound and complete local robustness analysis by encoding this task as a MIP problem. 
A classifier $D$ is locally robust at a neighborhood of an input $x$ if all inputs in the neighborhood are classified as some class $c'$. \citet{g_ref_33}'s verifier gets as arguments a classifier $D$, an input $x$ classified as $c'$, and a neighborhood $\mathcal{N}(x)$ defined by a series of intervals bounding each input entry, i.e., $x'\in \mathcal{N}(x)$ if and only if $\forall k.\ x'_k\in [l_k, u_k]$, where $k$ is the $k^\text{th}$ input entry. The verifier generates a MIP for every target class $c_t\neq c'$. 
If there is no solution to the MIP of $c_t$, the score of $c_t$ is lower than the score of $c'$ for all inputs in $\mathcal{N}(x)$; otherwise, there is an input whose score of $c_t$ is greater or equal to the score of $c'$. 
If no MIP has a solution, the network is locally robust at $\mathcal{N}(x)$. 
The verifier encodes the local robustness analysis as follows.
It introduces a real-valued variable for every input neuron, denoted $z_{0,k}$. The input neurons are subjected to the neighborhood's bounds: $\forall k.\ l_k \leq z_{0,k} \leq  u_k$. 
Each internal neuron has two real-valued variables $\hat{z}_{m,k}$, for the affine computation, and ${z}_{m,k}$, for the ReLU computation. 
The affine computation is straightforward: 
$\hat{z}_{m,k}=b_{m,k}+\sum_{k'=1}^{k_{m-1}}{w}_{m,k,k'}\cdot{z}_{m-1,k'}$. 
For the ReLU computation, the verifier introduces a boolean variable $a_{m,k}$ and two concrete (real-valued) lower and upper bounds $l_{m,k},u_{m,k}\in \mathbb{R}$, bounding the possible values of the input $\hat{z}_{m,k}$. The verifier encodes ReLU by the constraints: ${z}_{m,k}\geq0$, ${z}_{m,k}\geq \hat{z}_{m,k}$, ${z}_{m,k} \leq u_{m,k}\cdot a_{m,k}$, and ${z}_{m,k} \leq \hat{z}_{m,k}-l_{m,k}(1-a_{m,k})$. The concrete bounds are computed by solving the optimization problems $l_{m,k}=\min z_{m,k}$ and $u_{m,k}=\max z_{m,k}$ where the constraints are the encodings of all previous layers. 
We denote the set of constraints pertaining to the neurons by $\mathcal{D}$.
To determine local robustness, the verifier duplicates the above constraints $|C|-1$ times, once for every class $c_t\neq c'$.
For every $c_t$, the verifier adds the constraint 
$z_{L,c_t}-z_{L,c'}\geq 0$ where $L$ is the output layer. 
This constraint checks whether there is an input in the neighborhood 
whose $c_t$'s score 
is greater or equal to $c'$'s score. 
If the constraints of the MIP of $c_t$ are satisfied, the neighborhood has an input not classified as $c'$ and hence it is not locally robust.
Otherwise, no input in the neighborhood is classified as $c_t$. Overall, if at least one of the $|C|-1$ MIPs is satisfied, the neighborhood is not locally robust; otherwise, it is locally robust. 

\paragraph{Adaptations} \tool builds on the above encoding to compute the maximal globally non-robust bound. We next describe our adaptations supporting (1)~analysis of global robustness over any input and its perturbed example, 
(2)~encoding dependencies between the MIP variables,  
 and (3)~encoding the globally non-robust bound $\delta_{c',c_t}$ and its lower bound $\delta^{HA}_{c',c_t}$.
 
To support global robustness analysis, \tool has to consider any input and its perturbed example. Thus, \tool has two copies of the above variables and constraints. The first copy is used for the input, while the second copy is used for the perturbed example, and its variables are denoted with a superscript $p$. We denote the set of constraints of the network propagating the perturbed example by $\mathcal{D}^p$. To capture $x^p=f_P(x,\mathcal{E'})$, \tool adds an input constraint $\phi_{in}$ (defined in \Cref{sec:In_dep}).
\tool encodes $\mathcal{E'} \in I_{\mathcal{E}}$, limiting the amount of perturbation, by a constraint bounding each entry of $\mathcal{E}'$ in its interval in $I_\mathcal{E}$.
Additionally, \tool computes dependencies between the MIP variables. 
These dependencies emerge from the input dependencies, the perturbation range $I_{\mathcal{E}}$, and the network's computation. 
The dependencies enable \tool to reduce the problem's complexity without affecting the soundness and completeness of the encoding.
The dependency constraint $\phi_{dep}$ involves running additional MIPs (described in~\Cref{sec:semantic_perturbations}).

To encode the maximization of the globally non-robust bound $\delta$, \tool performs the following adaptations.
First, for every target class $c_t$, it introduces a variable $\delta_{c',c_t}$ capturing the maximal globally non-robust bound targeting $c_t$. Namely, it is the maximal class confidence of $c'$ over all inputs that have a perturbed example that is misclassified as $c_t$. Accordingly, it adds as an objective: $\max \delta_{c',c_t}$. 
Note that by adding this objective, \tool can solve the MIP in an anytime manner, because it looks for a number (the bound) and not a yes/no answer (like local robustness). 
Second, it encodes the constraint $\mathcal{C}(x,c',D)\geq \delta$ by replacing the constraint 
$z_{L,c_t}-z_{L,c'}\geq 0$ with constraints requiring that the class confidence of $c'$ is at least $\delta_{c',c_t}$: $\forall c''\in C\setminus\{c'\}.\ z_{L,c'}-z_{L,c''}\geq\delta_{c',c_t}$. 
Third, it encodes the constraint $\mathcal{C}(f_P(x,\mathcal{E'}),c_t,D)>0$ (the targeted version of the constraint $\mathcal{C}(f_P(x,\mathcal{E'}),c',D)\leq 0$).
This constraint is captured by the input constraint $\phi_{in}$ and the constraints: $\forall c''\in C\setminus\{c_t\}.\ z^p_{L,c_t}-z^p_{L,c''}>0$ (instead of the constraint $z_{L,c_t}-z_{L,c'}\geq 0$). In case $C_t=C\setminus \{c'\}$, 
the last constraints are defined over $\geq$ instead of $>$. 
Lastly, \tool adds the constraint $\delta_{c',c_t}\geq \delta_{c',c_t}^{HA}$, where $\delta_{c',c_t}^{HA}$ is the lower bound computed by the hyper-adversarial attack. 
To conclude, \tool submits to the MIP solver the following maximization problem for each $c_t$:
\begin{equation}\label{main_problem}
  \begin{gathered}
     \max \delta_{c',c_t} \hspace{0.5cm}\text{subject to}\\
     \mathcal{E'}\in I_\mathcal{E} ;\hspace{0.27cm} \phi_{in};\hspace{0.27cm} \phi_{dep}; \hspace{0.27cm}\delta_{c',c_t}\geq \delta^{HA}_{c',c_t};\hspace{0.27cm}\forall {c''\neq c'}.\ z_{L,c'}-z_{L,c''}\geq\delta_{c',c_t};\hspace{0.27cm}\forall c''\neq c_t.\ z^p_{L,c_t}-z^p_{L,c''}>0\\
     \forall m \forall k:\hspace{0.5cm}\;\hat{z}_{m,k}=b_{m,k}+\sum_{k'=1}^{k_{m-1}}{w}_{m,k,k'}\cdot{z}_{m-1,k'}; \hspace{0.5cm}\hat{z}^p_{m,k}=b_{m,k}+\sum_{k'=1}^{k_{m-1}}{w}_{m,k,k'}\cdot{z^p}_{m-1,k'}\\
      {z}_{m,k}\geq0; \hspace{0.5cm} {z}_{m,k}\geq \hat{z}_{m,k}; \hspace{0.5cm} {z}_{m,k} \leq u_{m,k}\cdot a_{m,k}; \hspace{0.5cm}\;{z}_{m,k} \leq \hat{z}_{m,k}-l_{m,k}(1-a_{m,k})\\
     {z^p}_{m,k}\geq0;\hspace{0.5cm} {z^p}_{m,k}\geq \hat{z}^p_{m,k};\hspace{0.5cm}
    {z^p}_{m,k} \leq u^p_{m,k}\cdot a^p_{m,k};\hspace{0.5cm}{z^p}_{m,k} \leq \hat{z}^p_{m,k}-l^p_{m,k}(1-a^p_{m,k})\\
    \end{gathered}
\end{equation}
From the above encoding, we get the next two lemmas.
\begin{lemma}
An optimal solution to the above MIP $\delta^*_{c',c_t}$ is the maximal globally non-robust bound targeting $c_t$.  
Consequently, $\delta^*=\max_{c_t\in C_t}\delta^*_{c',c_t}$ is the maximal globally non-robust bound of $C_t$.
\end{lemma}
\begin{lemma}
The minimal globally robust bound is $\delta=\delta^*+\Delta$, where $\Delta>0$ is the precision level.  
\end{lemma}
\subsection{Leveraging Dependencies to Reduce the Complexity}
\label{sec:semantic_perturbations}
In this section, we explain how to identify dependencies between the MIP's variables that enable to reduce the complexity of the MIP problem. \tool incorporates the dependencies via the constraints $\phi_{in}$, for dependencies between the input and its perturbed example, and $\phi_{dep}$ for dependencies between neurons of the two network copies. 
Input dependencies are defined by the perturbation. 
Neuron dependencies are computed by \tool from: the neurons' concrete bounds, dependency propagation from preceding layers, or MIPs.
We focus on input constraints $\phi_{in}$ defined by linear or quadratic constraints 
and dependency constraints $\phi_{dep}$ defined by equalities or inequalities over the MIP variables, all are supported by standard MIP solvers. 

\subsubsection{Dependency Constraint}
We begin with describing the dependency constraint $\phi_{dep}$. The goal is to identify pairs of neurons $z_{m,k}$ and $z^p_{m',k'}$, one from the input's network and the other one from the perturbed example's network, that satisfy equality or inequality constraint:   $z_{m,k}\bowtie z^p_{m',k'}$, where $\bowtie \in \{\leq, <, \geq, >, =\}$. These constraints prune the search space. To determine whether $z_{m,k}\bowtie z^p_{m',k'}$, for $\bowtie \in \{\leq, <, \geq, >, =\}$, 
one can solve 
a maximization and a minimization problems, each is over the constraints of Problem~\ref{main_problem} up to layers $m$ and $m'$, $\phi_{in}$, and the current $\phi_{dep}$ where the objectives are:
   \begin{equation*}
\begin{multlined}
\phantom{aa}u_{m,k,m',k'}=\text{max }\:{z_{m,k}-z^p_{m',k'}}\;\;\;
 l_{m,k,m',k'}=\text{min}\:{z_{m,k}-z^p_{m',k'}}
\end{multlined}
\end{equation*}
The values of $u_{m,k,m',k'}$ and $l_{m,k,m',k'}$ determine the relation between $z_{m,k}$ and $z^p_{m',k'}$ as follows:
\begin{lemma}
\label{lemma::secdep1}
 Given two variables, $z_{m,k}$ and $z^p_{m',k'}$ and their boolean variables $a_{m,k}$ and $a^p_{m',k'}$:
  \begin{itemize}[nosep,nolistsep]
  \item if $u_{m,k,m',k'}=l_{m,k,m',k'}=0$, then $z_{m,k} = z^p_{m',k'}$ and $a_{m,k} = a^p_{m',k'}$.
  \item if $l_{m,k,m',k'}\bowtie0$, then $z_{m,k} \bowtie z^p_{m',k'}$ and $a_{m,k} \geq a^p_{m',k'}$, where $\bowtie\in\{\geq,>\}$. 
  \item if $u_{m,k,m',k'}\bowtie0$, then $z_{m,k} \bowtie z^p_{m',k'}$ and $a_{m,k} \leq a^p_{m',k'}$, where $\bowtie\in\{\leq,<\}$.
  \end{itemize}
\end{lemma}

The proof follows from the MIPs and the monotonicity of ReLU. 
Solving these maximization and minimization problems for all combinations of $(m,k)$ and $(m',k')$ would significantly increase the complexity of \tool and is thus impractical. 
To cope, \tool focuses on two kinds of dependencies: those stemming directly from the perturbation and those stemming from the computation of corresponding neurons (i.e., $m'=m$ and $k'=k$). For the first kind of dependencies, \tool adds the input dependencies (\Cref{sec:In_dep}) and employs dependency propagation (defined shortly), neither requires solving the above optimization problems. 
For the second kind of dependencies, \tool first checks the concrete bounds: if the lower bound of $z_{m,k}$ is greater than the upper bound of $z_{m,k}^p$, then $z_{m,k}>z^p_{m,k}$ and vice-versa. Otherwise, \tool computes the dependencies by solving the minimization and maximization problems (only for corresponding neuron pairs). 
%
 

We next define dependency propagation. A dependency is propagated for a pair of neurons in the same layer $z_{m,k}$ and $z^p_{m,k'}$ if their inputs share dependencies that are preserved or reversed for $z_{m,k}$ and $z^p_{m,k'}$ because of their linear transformations and the ReLU definition:

\begin{lemma}[Dependency Propagation]\label{lem:depprop}
Given a neuron $z_{m,k}$ 
and a neuron $z^p_{m,k'}$ 
then: 
  \begin{itemize}[nosep,nolistsep]
  
  \item if~~$\forall i\ (w_{m,k,k_i}\cdot z_{m-1,k_i}=w_{m,k',k'_i}\cdot z^p_{m-1,k'_i})$, then $z_{m,k} = z^p_{m,k'}$ and $a_{m,k} = a^p_{m,k'}$.
  \item  if~~$\forall i\ ( w_{m,k,k_i}\cdot z_{m-1,k_i} \geq w_{m,k',k'_i}\cdot z^p_{m-1,k'_i}) $, then  ${z}_{m,k} \geq  {z}^p_{m,k'}$ and ${a}_{m,k} \geq {a}^p_{m,k'}$.
  \item  if~~$\forall i\ (w_{m,k,k_i}\cdot z_{m-1,k_i} \leq w_{m,k',k'_i}\cdot z^p_{m-1,k'_i}) $, then  ${z}_{m,k} \leq {z}^p_{m,k'}$ and ${a}_{m,k} \leq {a}^p_{m,k'}$.  
  


  \end{itemize}
  \end{lemma} 
\tool generally uses the lemma for $k'=k$, except for geometric perturbations (translation and rotation). In this case, 
it uses the lemma for indices whose correspondence stems from the perturbation, e.g., $k'=k+1$ for a translation perturbation that moves the pixels by one coordinate.

\paragraph{Example} 
We next exemplify how \tool uses this lemma. 
Consider the small network and the occlusion perturbation blackening the pixel $(1,1)$, presented in~\Cref{fig::conv_and_fully_global}. 
 \tool sets the dependencies in a matrix, called $\phi_{dep}$. An entry $(m,k)$ corresponds to the dependency constraint of the neuron $z_{m,k}$, where the entries at $(0,k)$ store the dependencies of the input neurons and index pairs $(i,j)\in[d_1]\times[d_2]$ are transformed to a single index: $k=(i-1)\cdot d_2+j\in [d_1\cdot d_2]$. Because the occlusion perturbation is not geometric, we focus on the dependencies of respective neurons $k=k'$. 
In this case, the weights are equal and \Cref{lem:depprop} is simplified to the following cases: 
  \begin{itemize}[nosep,nolistsep]
  \item if~~$\forall i.\  w_{m,k,k_i}\cdot(z_{m-1,k_i}-z^p_{m-1,k_i}) = 0$, then $z_{m,k} = z^p_{m,k}$ and $a_{m,k} = a^p_{m,k}$.
  \item  if~~$\forall i.\ w_{m,k,k_i}\cdot(z_{m-1,k_i}-z^p_{m-1,k_i})  \geq 0 $, then  ${z}_{m,k} \geq  {z}^p_{m,k}$ and ${a}_{m,k} \geq {a}^p_{m,k}$.
  \item  if~~$\forall i.\ w_{m,k,k_i}\cdot(z_{m-1,k_i}-z^p_{m-1,k_i})  \leq 0  $, then  ${z}_{m,k} \leq {z}^p_{m,k}$ and ${a}_{m,k} \leq {a}^p_{m,k}$.
  \end{itemize}
\tool first encodes the dependencies stemming from the occlusion perturbation (defined in~\Cref{sec:In_dep}):
\begin{itemize}[nosep,nolistsep]
  \item $\phi_{dep}[0,1]=\{z_{0,1}\geq z^p_{0,1}\}$, since $z^p_{0,1}=0$ and $z_{0,1}\in[0,1]$, and
  \item $\forall k\in[2,16].\ \phi_{dep}[0,k]=\{z_{0,k}=z^p_{0,k}\}$, since the perturbation does not change these pixels.
\end{itemize}
%
These dependencies are propagated to the first layer as follows:
\begin{itemize}[nosep,nolistsep]
  \item $\phi_{dep}[1,1]=\{z_{1,1}\geq z_{1,1}^p\}$: This constraint, corresponding to the neuron $z_{1,1}$, follows from the second bullet of \Cref{lem:depprop}. To check that the second bullet holds, \tool inspects the dependency constraints of the inputs to $z_{1,1}$ which are $z_{0,1},z_{0,2},z_{0,5},z_{0,6}$. It observes that $\phi_{dep}[0,1]=\{z_{0,1}\geq z^p_{0,1}\}$ and  $w_{1,1,1}=1\geq 0$, and  $\phi_{dep}[0,k]=\ \{z_{1,k}= z_{1,k}^p\}$, for $k\in\{2,5,6\}$. 
  \item $\phi_{dep}[1,k] = \{z_{1,k}= z_{1,k}^p\}$, for $k\in\{2,3,4\}$: These constraints, corresponding to the neurons $z_{1,2}$, $z_{1,3}$, and $z_{1,4}$, follow from the first bullet of \Cref{lem:depprop}. To check that the first bullet holds, \tool inspects the dependency constraints of the inputs to each of these neurons and observes that all have an equality constraint. 
\end{itemize}
These dependencies are propagated to the second layer as follows:
\begin{itemize}[nosep,nolistsep]
  \item $\phi_{dep}[2,1]=\{z_{2,1}\geq z_{2,1}^p\}$: This constraint, corresponding to the neuron $z_{2,1}$, follows from the second bullet of \Cref{lem:depprop}, since $\phi_{dep}[1,1]=\{z_{1,1}\geq z^p_{1,1}\}$, the weight $w_{2,1,1}=1$ is positive, and the dependencies of the other inputs are $\phi_{dep}[1,k]=\{z_{1,k}= z^p_{1,k}\}$, for $k>1$.
  \item $\phi_{dep}[2,2] =\{z_{2,2}\leq z^p_{2,2}\}$: This constraint, corresponding to the neuron $z_{2,2}$, follows from the third bullet of \Cref{lem:depprop}, since $\phi_{dep}[1,1]=\{z_{1,1}\geq z^p_{1,1}\}$, the weight $w_{2,1,2}=-1$ is negative, and the dependencies of the other inputs are equalities. 
\end{itemize}

\begin{algorithm}[t]
\DontPrintSemicolon
\KwIn{A perturbation $f_P$, its range $I_\mathcal{E}$, an input constraint $\phi_{in}$, the network encodings $\mathcal{D}$, $\mathcal{D}^p$.}
\KwOut{A dependency constraint $\phi_{dep}$.}

$\phi_{dep}[L+1,\max\{k_1,\ldots,k_L\}] = \bot$\;\label{ln:init}
$\phi_{dep}$ = add\_perturbation\_dep($f_P,I_\mathcal{E}$, $\phi_{in}$, $\mathcal{D}$, $\mathcal{D}^p$, $\phi_{dep}$)\;\label{ln:depinit}
\ForEach {\text{layer}$\;m \in \{1,\dots,L\} $}{\label{ln:layer}
    \ForEach {\text{neuron}$\;k \in \{1,\dots,k_m\}$}{\label{ln:neuron}
   \lIf{$\phi_{dep}[m,k]\neq\bot$}{\label{ln:skip}
      continue
   }
   \lIf{$l_{m,k}>u^p_{m,k}$}{\label{ln:concbeg}
      $\phi_{dep}[m,k]= \{z_{m,k}>z^p_{m,k}, a_{m,k} \geq a^p_{m,k}\}$
    }   
   \lElseIf{$u_{m,k}<l^p_{m,k}$}{\label{ln:concend}
      $\phi_{dep}[m,k]= \{z_{m,k}<z^p_{m,k}, a_{m,k} \leq a^p_{m,k}\}$
    }
    \Else{     
   $\phi_{dep}[m,k]$ = dependencies\_propagation($m$, $k$, $\mathcal{D}$, $\mathcal{D}^p$, $\phi_{dep}$)\;\label{ln:dep_prop}
   \lIf{$\phi_{dep}[m,k]\neq\bot$}{
      continue \label{ln:skip2}
   }
   $u_{m,k,m,k}=\infty$ , $l_{m,k,m,k}=-\infty$\; 
   \If{$[l_{m,k},u_{m,k}] \sqcap [l^p_{m,k},u^p_{m,k}]==[l_{m,k}, u^p_{m,k}]$
       }{\label{ln:minbeg}
          
        $l_{m,k,m,k} = \bold{min_{\text{cutoff}=0}}\: \: z_{m,k} -z^p_{m,k}\:s.t. \:\mathcal{E'} \in I_\mathcal{E} , \phi_{in}, \phi_{dep}, \mathcal{D}_{1:m}, \mathcal{D}^p_{1:m}$\label{ln:minend}
   } 
   \If{$[l_{m,k},u_{m,k}] \sqcap [l^p_{m,k},u^p_{m,k}]==[l^p_{m,k}, u_{m,k}]$
       }{\label{ln:maxbeg}
          
    $u_{m,k,m,k} = \bold{max_{\text{cutoff}=0}}\: \:z_{m,k} -z^p_{m,k}\:s.t. \:\mathcal{E'} \in  I_\mathcal{E} , \phi_{in}, \phi_{dep}, \mathcal{D}_{1:m}, \mathcal{D}^p_{1:m}$\label{ln:maxend}
   } 

   \lIf{($[l_{m,k,m,k},u_{m,k,m,k}]==[0,0]$)}{\label{ln:upbeg}
      $\phi_{dep}[m,k]=  \{z_{m,k}=z^p_{m,k}, a_{m,k}=a^p_{m,k}\}$
    }
   \lElseIf{$l_{m,k,m,k}\geq 0$}{
      $\phi_{dep}[m,k]= \{z_{m,k}\geq z^p_{m,k}, a_{m,k} \geq a^p_{m,k}\}$
    }   
   \lElseIf{$u_{m,k,m,k}\leq 0$}{
      $\phi_{dep}[m,k]= \{z_{m,k}\leq z^p_{m,k}, a_{m,k} \leq a^p_{m,k}\}$\label{ln:upend}
    }
}
 }
\lIf{ $\forall k.\:\phi_{dep}[m,k] ==\bot$} {break\label{ln:earlyter}} 
 }
\Return{$\bigwedge\{\phi_{dep}[m,k]\:|\:\phi_{dep}[m,k]\neq \bot, m>0\}$}\label{ln:return}
  \caption{Compute\_dependency\_constraints($f_P$, $I_\mathcal{E}$, $\phi_{in}$, $\mathcal{D}$, $\mathcal{D}^p$)}\label{alg:dependencies}
\end{algorithm}

\paragraph{Algorithm}
\Cref{alg:dependencies} shows how \tool computes the dependencies. 
Its inputs are the perturbation function $f_P$, its range $I_\mathcal{E}$, the input constraint $\phi_{in}$, and the classifier encodings $\mathcal{D}$ and $\mathcal{D}^p$. 
It begins by initializing a matrix, whose rows are the layers (including the input layer) and its columns are the neurons in each layer. An entry $(m,k)$ in the matrix corresponds to the dependency constraint of the neuron $z_{m,k}$ (\Cref{ln:init}). 
\tool begins by adding the perturbation's dependencies for the relevant neurons, as defined in~\Cref{sec:In_dep} (\Cref{ln:depinit}). 
\tool then iterates over all neurons, layer-by-layer, and computes dependencies (\Cref{ln:layer}--\Cref{ln:neuron}). 
If an entry $(m,k)$ is not empty, it is skipped (\Cref{ln:skip}).  
Otherwise, if $l_{m,k}$, the lower bound of $z_{m,k}$, is greater than $u_{m,k}^p$, the upper bound of $z^p_{m,k}$, 
then $z_{m,k}>z^p_{m,k}$ and vice-versa (\Cref{ln:concbeg}--\Cref{ln:concend}).
Then, \tool checks whether it can propagate a dependency (\Cref{ln:dep_prop}). 
If not,
\tool checks whether it should solve the optimization problems. 
The minimization problem is solved if $l^p_{m,k} \leq l_{m,k}\leq u^p_{m,k} \leq u_{m,k}$, in which case it may be that $z_{m,k}\geq z^p_{m,k}$ (\Cref{ln:minbeg}--\Cref{ln:minend}). 
Similarly, 
the maximization problem is solved if 
$l_{m,k} \leq l^p_{m,k}\leq u_{m,k} \leq u^p_{m,k}$, in which case it may be that $z^p_{m,k}\geq z_{m,k}$  (\Cref{ln:maxbeg}--\Cref{ln:maxend}).      
Note that if $l^p_{m,k} < l_{m,k}\leq u_{m,k} < u^p_{m,k}$, the two variables are incomparable since $z^p_{m,k}$ may be equal to $l^p_{m,k}$ (and thus $z^p_{m,k}< z_{m,k}$) or equal to $u^p_{m,k}$ (and thus $z_{m,k}< z^p_{m,k}$). Similarly, the two variables are incomparable if $l_{m,k} < l^p_{m,k}\leq u^p_{m,k} < u_{m,k}$.
The minimization problem early stops when the lower bound on the objective reaches zero or when the upper bound becomes negative. The maximization problem early stops when the upper bound on the objective reaches zero or when the lower bound becomes positive. This is because the optimal minimal and maximal values are not required to determine the relation between the variables. 
If both returned values are zero, the variables are equal. Otherwise, if the minimal value is at least zero, then
$z_{m,k}\geq z^p_{m,k}$, and if the maximal value is at most zero, then $z_{m,k}\leq z^p_{m,k}$ (\Cref{ln:upbeg}--\Cref{ln:upend}).
The boolean variables have a similar dependency.
In case no pair of neurons depends in the current layer, the algorithm terminates 
(\Cref{ln:earlyter}). Finally, the algorithm returns all dependencies (\Cref{ln:return}).

\subsubsection{Input Dependencies}\label{sec:In_dep}
Lastly, we describe the input dependencies $\phi_{in}$ and define the dependencies' encoding 
 for several semantic perturbations and the $L_\infty$ perturbation. These perturbations have been shown to inflict adversarial examples~\cite{ref35,g_ref_13,g_ref_14,g_ref_15,g_ref_16,g_ref_17,g_ref_18}. 
Our definitions assume grayscale images, for simplicity's sake, but \tool supports colored images. 
To align with our MIP encoding, we transform index pairs $(i,j)\in[d_1]\times[d_2]$ to a single index: $k=(i-1)\cdot d_2+j\in [d_1\cdot d_2]$.

\paragraph{Brightness$([l,u])$} Brightness, whose range is an interval $-1\leq l \leq u \leq 1$, adds a value $\mathcal{E'} \in [l,u]$ to the input $x$. Namely, 
$\phi_{in}= \forall k.\ ( z^p_{0,k}= z_{0,k}+\mathcal{E'})$. 
To enhance the encoding's accuracy, \tool adds constraints depending on the sign of $\mathcal{E'}$.
Given the sign of $\mathcal{E'}$,
\tool computes the sign of every neuron $z_{1,k}$ in the first layer, $s_k=sign(\mathcal{E'}\sum_{i'} w_{1,k,i'})$, and adds $\phi_{dep}[1,k]=\{{z}_{1,k} \bowtie {z}^p_{1,k},{a}_{1,k} \bowtie {a}^p_{1,k}\}$, where $\bowtie\triangleq\leq$ if $s_k= +$, $\bowtie\triangleq\geq$ if $s_k=-$, and $\bowtie\triangleq =$ if $s_k=0$.
 The sign of $\mathcal{E'}$ is $-$ if $u\leq 0$ and is $+$ if $l\geq 0$. Otherwise, \tool duplicates the encoding and handles the cases $\mathcal{E'}\in [l,0]$ and $\mathcal{E'}\in [0,u]$ separately.
\tool propagates dependencies to subsequent layers as defined in~\Cref{lem:depprop}.

\paragraph{Contrast$([l,u])$}  Contrast, whose range is an interval $l \leq u \in \mathbb{R^{+}}$, multiplies the input by a value $\mathcal{E'} \in [l,u]$. Namely, $\phi_{in}=\forall k.\ (z^p_{0,k}= \mathcal{E'}\cdot z_{0,k})$.
To propagate dependencies as in~\Cref{lem:depprop}, \tool also adds the constraints $\forall k. \ \phi_{dep}[0,k]=\{{z_{0,k} \bowtie z^p_{0,k}}\}$, where $\bowtie\triangleq \leq$ if $\mathcal{E'} \geq 1$, and $\bowtie\triangleq \geq$ if $\mathcal{E'} \leq 1$. The condition $\mathcal{E'} \geq 1$ is true if $l\geq 1$; the condition $\mathcal{E'} \leq 1$ is true if $u\leq 1$; otherwise, \tool duplicates the encoding and handles the cases $\mathcal{E'}\in [l,1]$ and $\mathcal{E'}\in [1,u]$ separately.


\paragraph{Occlusion$([i_l,i_u],[j_l,j_u],[w_l,w_u])$} Occlusion's range is three intervals $1\leq i_l \leq i_u \leq d_1$; $1\leq j_l \leq j_u \leq d_2$ and $1\leq w_l \leq w_u \leq min(d_1-i_u,d_2-j_u)$. It perturbs based on a triple $(i, j, w)$, where $(i,j)$ defines the top-left coordinate of the occlusion square $S$ and $w$ is its length. 
The perturbation function maps every pixel in the square to zero; other pixels remain as are, namely $\phi_{in}=\forall k.\ (k\in S\Rightarrow z^p_{0,k}= 0)\wedge(k\notin S\Rightarrow z^p_{0,k}= z_{0,k}) $. 
To propagate dependencies as in~\Cref{lem:depprop}, \tool also adds the constraints $\forall k\in S.\ \phi_{dep}[0,k] =\{z_{0,k} \geq z^p_{0,k}\}$ and $\forall k\notin S.\ \phi_{dep}[0,k] =\{z_{0,k} = z^p_{0,k}\}$. 

\paragraph{Patch$([0,\epsilon],[i_l,i_u],[j_l,j_u],[w_l,w_u])$} Patch's range consists of a perturbation limit $\epsilon\in (0,1]$
and the three intervals as defined in the occlusion's range. A patch is defined by a top-left coordinate $(i,j)$ of the patch's square $S$ and the patch length $w$. 
Its function, generalizing occlusion, perturbs every pixel in the patch by a value in the interval $[-\epsilon,\epsilon]$ (unlike occlusion which maps to zero); other pixels remain as are, namely: 
 $\phi_{in}=\forall k.\ ({k\in S}\Rightarrow\ z^p_{0,k} \leq z_{0,k}+\epsilon
 \wedge z^p_{0,k} \geq z_{0,k}-\epsilon) \wedge (k\notin S\Rightarrow z^p_{0,k}=z_{0,k})$. Note that since every pixel in $S$ can be perturbed or keep its original value, a patch defined by $(i,j)$ and $w$ also contains every smaller, subsumed patch. 
To propagate dependencies as in~\Cref{lem:depprop}, \tool also adds the constraints $\forall k\notin S.\ \phi_{dep}[0,k] =\{z_{0,k} = z^p_{0,k}\}$.

\paragraph{Translation$([t_x^l,t_x^u],[t_y^l,t_y^u])$} Translation's range is two intervals $-d_1\leq t_x^l\leq t_x^u\leq d_1$ and $-d_2\leq t_y^l\leq t_y^u\leq d_2$. Given a coordinate $(t_x,t_y)$, 
the perturbation function moves every pixel by $(t_x,t_y)$, namely $\phi_{in}=\forall(i,j).\ (i-t_x\notin [d_1]\lor j-t_y \notin[d_2]\Rightarrow z^p_{0,k}=0)\wedge (i- t_x\in[d_1] \wedge j-t_y\in[d_2] \Rightarrow z^p_{0,k}=z_{0,k-t_x\cdot d_2-t_y})$. 
To propagate dependencies as in~\Cref{lem:depprop}, \tool also adds a constraint for every $k=(i-1)\cdot d_2+j$: if $(i-t_x\notin [d_1]\lor j-t_y \notin[d_2])$ then $\phi_{dep}[0,k] =\{z_{0,k} \geq z^p_{0,k}\}$, otherwise $\phi_{dep}[0,k] =\{ z_{0,k-t_x\cdot d_2-t_y}=z^p_{0,k}\}$. 

\paragraph{Rotation$([\theta_l,\theta_u])$} Rotation, whose range is an interval $0\leq \theta_l\leq \theta_u\leq 360$, rotates the input by an angle $\theta$ using a standard rotation algorithm. 
The algorithm\ifthenelse{\boolean{is_conference}}{\footnote{The pseudo-code is provided in the extended version of this paper~\cite[Appendix A]{g_ref_81}.}}{~(\Cref{sec:appex_rot})} begins by centralizing the pixel coordinates, i.e., a coordinate $(i, j)$ is mapped to $i_c = i - d_1/2$ and $j_c = j - d_2/2$. The centralized coordinates are rotated by $i_r=i_c \cdot\sin(\frac{\theta \cdot \pi}{180}) + j_c\cdot \cos(\frac{\theta \cdot \pi}{180})) + d_1/2$ and 
$j_r = (i_c \cdot \cos(\frac{\theta \cdot \pi}{180}) - j_c \sin(\frac{\theta \cdot \pi}{180}) + d_2/2$. Finally, a bilinear interpolation is executed.
Namely, $\phi_{in}=\forall(i,j).\ (1\leq i_r \leq d_1\wedge\;1\leq j_r \leq d_2\Rightarrow\:z^p_{0,k}=\text{bilinear\_interpolation}(z_{0,k},i_r,j_r))\wedge  
(i_r < 1\lor i_r >d_1\lor j_r < 1\lor j_r >d_2\Rightarrow z^p_{0,k}=0)$.
To propagate dependencies as in~\Cref{lem:depprop}, \tool also adds a constraint for every $k=(i-1)\cdot d_2+j$: 
if $i_r < 1\lor i_r >d_1\lor j_r < 1\lor j_r >d_2$, then $\phi_{dep}[0,k] =\{z_{0,k} \geq z^p_{0,k}\}$. 
We note that due to the complexity of the rotation algorithm, our implementation currently supports only $\theta_l=\theta_u$.

\paragraph{$L_{\infty}([0,\epsilon])$} $L_\infty$'s range is an interval $[0,\epsilon]$, where $\epsilon\in (0,1]$. It perturbs an input by changing every pixel by up to $|\epsilon|$. 
Namely, $\phi_{in}=\forall k.\ z^p_{0,k} \leq z_{0,k}+\epsilon
 \wedge z^p_{0,k} \geq z_{0,k}-\epsilon$.
 
\subsection{Suboptimal Lower Bounds and Hints via a Hyper-Adversarial Attack}
\label{sec:sub_opt_sol_and_opt_hints}

In this section, we explain how \tool computes suboptimal lower bounds to Problem~\ref{problem2} as well as optimization hints. 
For every target class $c_t\in C_t$, \tool computes a lower bound $\delta^{HA}_{c',c_t} \geq 0$, which is added as a constraint to the MIP problem corresponding to $c_t$, thereby pruning the search space. 
As part of this computation, \tool identifies \emph{hints}, 
guiding the MIP solver towards the maximal globally non-robust bound.  
The hints are provided to the MIP solver through two matrices $\mathcal{H}_{c',c_t}$, for the boolean variables in $\mathcal{D}$, and $\mathcal{H}^p_{c',c_t}$, for $\mathcal{D}^p$. An entry $(m,k)$ corresponds to the initial assignment of the boolean variable $a_{m,k}$ or $a^p_{m,k}$ and it is $0$, $1$, or $\bot$ (no assignment). 

\begin{algorithm}[t]
\DontPrintSemicolon
\KwIn{A classifier $D$, a class $c'$, a target class $c_t$, a perturbation $f_P$, its range $I_\mathcal{E}$, a dataset $DS$.}
\KwOut{A lower bound $\delta^{HA}_{c',c_t}$ and boolean hints $\mathcal{H}_{c',c_t}, \mathcal{H}_{c',c_t}^p \in\{0,1,\bot\}^{L\times \max\{k_1,\ldots,k_L\}}$.}

$\widetilde{DS}$ = sort($DS \cup random\_inputs(|DS|),\ \lambda x.\  \mathcal{C}(x,c',D))$ \label{alg2_ln:build_init_dataset}\;
$cands$ = $[x\in \widetilde{DS} \mid  \mathcal{C}(x,c',D)>0]$
\label{alg2_ln:classified_c}\;
$X$ = $|cands|>M$ ? $(cands[{1:\lfloor \frac{|cands|}{M} \rfloor: |cands|}])$ : $\widetilde{DS}[{1:M}]$\label{alg2_ln:sample_inputs}\;
$\widetilde{X}$ = $\{\tilde{x}_1,\ldots,\tilde{x}_M \}$; $\widetilde{\mathcal{E}}'$ = $\{\tilde{\mathcal{E}}'_1,\ldots,\tilde{\mathcal{E}}'_M \}$\label{alg2_ln:init_vars}\;
optimize($\max_{\widetilde{X},\widetilde{\mathcal{E}}'\in I_\mathcal{E}}\sum_{i=1}^M{\mathcal{C}(x_i+\tilde{x}_i,c',D)} + \sum_{i=1}^M{\left(\lambda_0 \cdot \frac{||\nabla_{\tilde{x}_i} 
\mathcal{C}(x_i+\tilde{x}_i,c',D)||}{||\nabla_{\tilde{x}_i} \min(\mathcal{C}(f_P(x_i+\tilde{x}_i,\tilde{\mathcal{E}}'_i),c_t,D),\tau)+\kappa||}\right) \cdot\min(\mathcal{C}(f_P(x_i+\tilde{x}_i,\tilde{\mathcal{E}}'_i),c_t,D),\tau)})$  \label{alg2_ln:opt}\;
sol = $\{x_i\mid \mathcal{C}(x_i+\tilde{x_i},c',D) >0\text{ and } \mathcal{C}(f_P(x_i+\tilde{x}_i,\tilde{\mathcal{E}}'_i),c_t,D) >0\}$
\label{alg2_ln:get_results_cc}\;

\If{$sol \neq \emptyset$}{
  $\delta^{HA}_{c',c_t}$ = $\max(\{\mathcal{C}(x_i+\tilde{x}_i,c',D)~\mid~x_i \in sol\})$ \label{alg2_ln:delta_s} \;
\For{$m\in [L]$ and $k\in [k_m]$}{
$active$ = $\frac{|\{D(x_i+\tilde{x}_i)_{m,k}\geq 0~\mid~x_i \in sol\}|}{|sol|}$; 
$active_p$ = $\frac{|\{D(f_P(x_i+\tilde{x}_i,\tilde{\mathcal{E}}'_i))_{m,k}\geq 0~\mid~ x_i \in sol\}|}{|sol|}$\label{alg2_ln:count}\; 
$\mathcal{H}_{c',c_t}[m,k]$ = $active>r$? 1 : ($1-active>r$? $0$ : $\bot$) \label{alg2_ln:h}\;
$\mathcal{H}^p_{c',c_t}[m,k]$ = $active_p>r$? 1 : ($1-active_p>r$? $0$ : $\bot$) \label{alg2_ln:hp}\;
}

\Return{$\delta^{HA}_{c',c_t}$, $\mathcal{H}_{c',c_t}$, $\mathcal{H}_{c',c_t}^p$}\label{alg2_ln:return}
    
}   
\lElse{\Return{0, $\bot^{L\times \max\{k_1,\ldots,k_L\}}$, $\bot^{L\times \max\{k_1,\ldots,k_L\}}$}\label{alg2_ln:return2}}
  \caption{Compute\_lower\_bound\_and\_hints($D$, $c'$, $c_t$, $f_P$, $I_\mathcal{E}$, $DS$)}\label{alg:sub_opt}
\end{algorithm}

The lower bound and hints are computed by a hyper-adversarial attack, as described in~\Cref{sec:key_idea_feasible_solutions}.
\Cref{alg:sub_opt} shows the computation. 
Its inputs are the classifier $D$, the class $c'$,  the target class $c_t$, the perturbation function $f_P$, its range $I_\mathcal{E}$, and an image dataset \texttt{DS} (e.g., the dataset used for training the network).  
\tool begins by expanding the dataset with \texttt{|DS|} random inputs (uniformly sampled over the range $[0,1]^{l\times d_1\times d_2}$), running each input through the classifier, and sorting them by their class confidence of $c'$ (\Cref{alg2_ln:build_init_dataset}). 
The image dataset is expanded in order to reach inputs with a broader range of class confidences. 
Accordingly, it identifies candidates for the hyper-input, \texttt{cands}, which are inputs classified as $c'$ (\Cref{alg2_ln:classified_c}).
Then, it constructs the hyper-input $X$, consisting of $M$ inputs (where $M$ is a hyper-parameter), as follows. 
If $|cands|$ is greater than $M$, \tool selects inputs from candidates in uniform steps of $\lfloor \frac{|cands|}{M}\rfloor$, from one to $|cands|$ (recall that \texttt{cands} is sorted by the class confidence). 
Otherwise, $X$ is the top-$M$ inputs from the expanded (sorted) dataset (\Cref{alg2_ln:sample_inputs}). 
It then initializes the input variables $\widetilde{X}$ and the perturbation variables $\widetilde{\mathcal{E}'}$ of Problem~\ref{problem3}. \tool adapts the loss described in~\Cref{sec:key_idea_feasible_solutions} to target $c_t$ and maximizes it (\Cref{alg2_ln:opt}). 
At high-level, the optimization begins from the hyper-input $X$ (since $\widetilde{X}$ is initially the zero vector), consisting of inputs that are classified as $c'$. The optimization has two objectives: (1) finding inputs $X+\widetilde{X}$ and perturbations $\mathcal{E}'$ leading to adversarial examples (i.e., the perturbed examples are classified as $c_t$) and 
(2)~maximizing the class confidence of the inputs in $X+\widetilde{X}$.
  These conflicting goals are balanced using an adaptive term, described shortly. To keep the conflict between the goals to the minimum required, the loss only requires that the perturbed examples' confidence in $c_t$ is positive. 
We note that \tool optimizes the loss simultaneously over $\widetilde{X}$ and $\widetilde{\mathcal{E}'}$. 
 The loss optimization is executed by projected gradient descent (PGD)~\cite{g_ref_6}.  
We next describe
\tool's adaptive balancing term $\lambda_i$. 
This term favors maximization of the class confidence of $c_t$, as long as $f_P(x_i+\tilde{x}_i,\tilde{\mathcal{E}}'_i)$ is not classified as $c_t$ (i.e., the confidence is not positive). 
Technically, 
this term multiplies a hyper-parameter $\lambda_0>1$ by the ratio of the gradient of the input's class confidence and the gradient of the perturbed example's class confidence. Multiplying by this ratio   
 enforces a proportion of $\lambda_0$ between the two optimization goals.
Namely, the smaller the second goal's gradient, the smaller this ratio's denominator, the larger the ratio and thus the larger the balancing term (which is multiplied by the second goal to keep the $\lambda_0$ proportion between the goals).
  Similarly, the smaller the first goal's gradient, the smaller this ratio's numerator, the smaller the ratio and thus the smaller the balancing term.
To stop maximizing the class confidence $c_t$ when it is positive, the denominator is the minimum of this confidence and a small number $\tau>0$. To avoid dividing by a number close to zero, the denominator is added a stability hyper-parameter $\kappa$. 
After the optimization, \tool retrieves the solutions, i.e., the inputs $x_i+\tilde{x}_i$ classified as $c'$ whose corresponding perturbed example is classified as $c_t$ (\Cref{alg2_ln:get_results_cc}).  
 The lower bound $\delta^{HA}_{c',c_t}$ is the maximal class confidence of the solutions (\Cref{alg2_ln:delta_s}). 
 To compute the hints, the solutions and their perturbed examples are run through the classifier.
 The hint $\mathcal{H}_{m,k}$ (or $\mathcal{H}^p_{m,k}$) of neuron $(m,k)$ is computed by first counting the number of solutions (or the perturbed examples) in which the neuron is active and dividing by the number of solutions (\Cref{alg2_ln:count}). 
  If this ratio is greater than a hyper-parameter $r$, the hint is $1$, if the complementary ratio is greater than $r$, the hint is $0$, otherwise it is $\bot$ (\Cref{alg2_ln:h}-\Cref{alg2_ln:hp}). 
 \tool returns the lower bound and the hints (\Cref{alg2_ln:return}).   
 If no solution is found, the trivial lower bound and empty hints are returned (\Cref{alg2_ln:return2}).

\subsection{Correctness and Running Time}
\label{sec:correctness_and_running_time}
In this section, we discuss correctness and running time analysis. 
\paragraph{Correctness} 
Our main theorem is that \tool is sound: it 
returns an interval containing the minimal globally robust bound. Given enough time, the bound is exact up to the precision level $\Delta$. 
\begin{restatable}[]{theorem}{ftc}
\label{thm1}
Given a classifier $D$, a class $c'$, a target class set $C_t$, and a perturbation $(f_P,I_\mathcal{E})$,  
\tool returns an interval
containing the minimal globally robust bound $\delta_{c',C_t}\in [\delta^l_{c',C_t},\delta^u_{c',C_t}]$. 
\end{restatable}
\begin{proof}[Proof Sketch]
 We show that (1)~$\delta^u_{c',C_t}$ is an upper bound on $\delta_{c',C_t}$ and (2)~$\delta^l_{c',C_t}$ is a lower bound, i.e., there is an input whose class confidence of $c'$ is $\delta^l_{c',C_t}-\Delta$ and it has an adversarial example classified as a class in $C_t$ obtained using the perturbation $(f_P,I_\mathcal{E})$. 
 We prove correctness for every $c_t\in C_t$ separately and because \tool returns the interval which is the maximum of the bounds, the claim follows. Let $c_t\in C_t$ and consider its MIP. 
The MIP soundly encodes Problem~\ref{problem2}: 
\begin{itemize}[nosep,nolistsep]
\item  The encoding of the network's computation is sound as proven by \citet{g_ref_33}.
\item The dependency constraints are sound: if a relation is added between two variables, it must exist. This follows from the perturbations' definitions and our lemmas. We note that due to scalability, \tool may not identify all dependency constraints, but it only affects the execution time of the anytime algorithm, not the MIP's soundness. 
\item The lower bound's constraint is sound since it is equal to the class confidence of an input satisfying Problem~\ref{problem2}'s constraint. 
\end{itemize}
By the anytime operation of the MIP solver, $\delta^l_{c',c_t}-\Delta$ is a solution to Problem~\ref{main_problem}'s constraints and thus $\delta^l_{c',c_t}$ is a lower bound.
Similarly, 
$\delta^u_{c',c_t}-\Delta$ is an upper bound for the maximally globally non-robust bound, and thus $\delta^u_{c',c_t}$ is an upper bound for the minimally globally robust bound.
  \end{proof}


\paragraph{Running time} The runtime of \tool is the sum of $T_e + T_{dep} + |C_t| \cdot T_{HA}  + |C_t| \cdot T_{\text{MIP}}$, 
where $T_e$ is the execution time of the MIP encoding (including the computation of the concrete bounds),    
$T_{dep}$ is the execution time of computing the dependencies $\phi_{dep}$,   
$T_{HA}$ is the execution time of the hyper-adversarial attack, 
and $T_{\text{MIP}}$ is the MIP solver's timeout for solving Problem~\ref{main_problem}. 
Our implementation reduces the running time by parallelizing the dependency computation on CPUs and parallelizing the hyper-adversarial attack on GPUs. 
The dominating factor is the execution time of the MIP solver given Problem~\ref{main_problem}. 
To mitigate it, our implementation parallelizes the $|C_t|$ MIPs over CPUs. 

\section{Evaluation}
\label{sec:eval}
In this section, we evaluate \tool. We begin by describing our experiment setup, then describe the baselines, and then present our experiments.

\paragraph{{Experiment setup}} We implemented \tool in Julia, as a module in
MIPVerify\footnote{https://github.com/vtjeng/MIPVerify.jl}~\cite{g_ref_33}.
The MIP solver is Gurobi 10.0 and the MIP solver timeout is three hours. 
The hyperparameters of \Cref{alg:sub_opt} are $M=10,000$, $r=0.95$ and $\lambda_0=1.01$. 
Experiments ran on an Ubuntu 20.04.1 OS on a dual AMD EPYC 7713 server with 2TB RAM and eight A100 GPUs. We evaluated \tool over several datasets.
MNIST~\cite{ref41} and Fashion-MNIST~\cite{ref58} consist of $28\times 28$ grayscale images, showing handwritten digits and clothing items, respectively. 
CIFAR-10~\cite{ref42} consists of  $3\times 32\times 32$ colored images, showing objects (e.g., a dog).
We consider several fully-connected and convolutional networks (\Cref{tab:models_table}).  
Their activation function is ReLU. The $3\times 50$ network is trained with PGD~\cite{g_ref_6}, an adversarial training defense. 
While the network sizes may look small compared to the network sizes analyzed by existing local robustness verifiers, we remind that \tool reasons about \emph{global robustness}. We consider network sizes that are an order of magnitude larger than those analyzed by Reluplex~\cite{g_ref_30}.

\begin{table}[t]
\small
\begin{center}
\caption{The networks used in our experiments.}
\begin{tabular}{lllcc}
    \toprule
Dataset & Name & Architecture & \#Neurons & Defense \\
\midrule
MNIST &  $3\times 10$ & 3 fully-connected layers & 20 &-\\
  &  $3\times 50$ & 3 fully-connected layers & 100 &PGD\\
  &  $conv1$ & 2 convolutional (stride 3) and 2 fully-connected layers & 550 &-\\
  &  $conv2$ & 2 convolutional (stride 1) and 2 fully-connected layers & 2077 &-\\
\midrule
Fashion-MNIST              &  $conv1$& 2 convolutional (stride 3) and 2 fully-connected layers & 550 &-\\
\midrule
CIFAR-10 &  $conv1$ & 2 convolutional (stride 3) and 2 fully-connected layers & 664 &-\\
\bottomrule
\end{tabular}
    \label{tab:models_table}
\quad
\end{center}
\end{table}
\begin{figure}[t]
    \centering
  \includegraphics[width=1\linewidth, trim=0 325 0 0, clip,page=8]{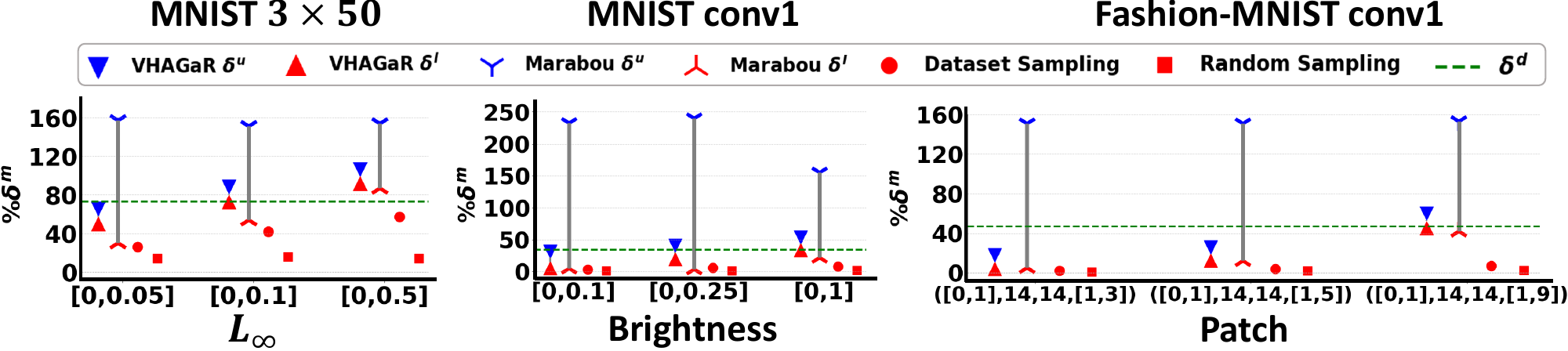}
    \caption{The upper and lower bounds of \tool vs. Marabou and sampling approaches.}
    \label{fig::demonstration}
\end{figure}
\paragraph{Baselines} 
We compare \tool to Marabou~\citep{g_ref_31}, a robustness verifier, and to sampling approaches~\citep{g_ref_41, g_ref_44}. 
Marabou, which improves Reluplex~\citep{g_ref_30}, is an SMT-based verifier that can determine whether a classifier is $\delta$-globally robust, for a given $\delta$ (although the paper focuses on local robustness tasks). We extend it to compute the minimal bound using binary search.
Since Marabou does not support quadratic constraints, we cannot evaluate it for the contrast perturbation.
Recently, a MIP-based verifier for analyzing global robustness of networks (not classifiers) has been proposed~\citep{g_ref_1, g_ref_2}. 
However, to date, their code is not available. 
There are many sampling approaches that estimate a global robustness bound by analyzing the local robustness of input samples~\citep{g_ref_41,g_ref_42,g_ref_43,g_ref_44}. These approaches either sample from a given dataset, e.g., DeepTRE~\citep{g_ref_41} that focuses on $L_0$ perturbations, or from the input domain, e.g., Groma~\citep{g_ref_44}. Groma computes the probability of an adversarial example with respect to $(\epsilon,\delta)$ global robustness in an input region $I$, i.e., the probability that inputs in $I$ whose distance is at most $\epsilon$ have outputs whose distance  is at most $\delta$: $\forall x, x'\in I.\ ||x-x'|| \leq \epsilon \Rightarrow ||D(x)-D(x')|| \leq \delta$.
Since these works focus on different global robustness definitions, we adapt their underlying ideas to our setting of computing the minimal global robustness bound, as we next describe. 
\emph{Dataset Sampling} computes the maximal globally non-robust bound by 
first estimating the maximal locally non-robust bound for every input in a given dataset of 70,000 samples. Then, it returns the maximal bound, denoted $\delta^{DS}$. 
\emph{Random Sampling} draws 70,000 independent and identically distributed (i.i.d.) samples from the input domain, assuming a Gaussian distribution, and computes the maximal locally non-robust bound for each. Like Groma, it returns the average non-robust bound $\delta^{RS}$ along with its 5\% confidence interval $h$ derived using the Hoeffding inequality~\citep{g_ref_67}, i.e., $\delta^{RS}\pm h$.  
Due to the very high number of samples, the baseline implementations cannot compute the maximal locally non-robust bound by running the MIP verifier. Instead, we take the following approach.
For discrete perturbations (e.g., occlusion), we perturb the input samples and run them through the classifier. We filter the perturbed examples that are misclassified and accordingly, we infer the maximal locally non-robust bound. For continuous perturbations (e.g., brightness), we run the verifier VeeP~\cite{g_ref_58}, designated for semantic perturbations. 
Note that the sampling baselines provide only a lower bound on the maximal globally non-robust bound. 
Nevertheless, our baseline implementations are challenging for \tool: the number of samples is 70,000, whereas DeepTRE is evaluated on at most 5,300 samples and Groma on 100 samples.


\begin{table*}[t]
\small
\begin{center}
\caption{\tool vs. Marabou and sampling approaches over MNIST classifiers. Sa. abbreviates Sampling, Occ. Occlusion, Pa. Patch, and Trans. Translation. NA is not applicable.}
\begin{tabular}{ll ccc cc cc ccc}
\toprule
Model & Perturbation             & \multicolumn{3}{c}{\tool} & \multicolumn{3}{c}{Marabou} & \multicolumn{2}{c}{ Dataset Sa.} & \multicolumn{2}{c}{Random Sa.}\\
\cmidrule(lr){3-5} \cmidrule(lr){6-8} \cmidrule(lr){9-10}  \cmidrule(lr){11-12} 
      &                          & $\delta^l$ &  $\delta^u$ &  $t$ & $\delta^l$ &  $\delta^u$ &  $t$ & $\delta^{DS}$ & $t$ & $\delta^{RS}\pm h$  & $t$  \\
      &                          & \%         &  \%         &   $[m]$     & \%         &  \%         &  $[m]$ & \%         &    $[m]$   & \%          & $[m]$ \\
\midrule
 MNIST            & Brightness([0,0.25])   & 28          & 28      & 0.1      & 24       &  114       & 161 & 8       & 0.5   & $1.7\pm0.2$   & 0.2    \\ 
$3\times 10$            & Brightness([0,0.1])    & 15          & 15      & 0.2      & 12       &  115       & 158 & 6       & 0.4   & $1.5\pm0.2$   & 0.2    \\ 
$\delta^m=45$   & Contrast([1,1.5])      & 0.4         & 0.8     & 180       & NA        &  NA         & NA   & 0.2     & 0.5   & $0.08\pm0.5$  & 0.7    \\
$\delta^d=29$\%  & Contrast([1,2])        & 0.6         & 1.6     & 180      & NA        &  NA         & NA   & 0.4     & 0.4   & $0.1\pm0.4$   & 0.7   \\
\midrule
MNIST            & $L_\infty$([0,0.1])    &  80         &    81   &   0.4   &   54      &   154     & 180 &   42    &  0.1  &  $6.1\pm10.6$   &  0.2  \\
$3\times 50$             & $L_\infty$([0,0.05])   &  57         &    58   &   3.9   &   30     &    157     & 180  &   26    &  0.2  &  $7.0\pm7.8$    &  0.3   \\
 $\delta^m=3.3$             & Brightness([0,0.25])          &  37         &    38   &   5.3    &  27      &    153     & 136 &   11    &  0.4  &  $4\pm0.6$      &  0.4  \\ 
 $\delta^d=73$\%  & Brightness([0,0.1])    &  22         &    22   &   9.2    &  15      &    154     & 141 &   6     &  0.4  &  $3\pm0.6$      &  0.4  \\
  & Occ.(14,14,9)         &  72         &    72   &   0.8  &  64      &    152     & 133 &   19    &  0.2  &  $3\pm0.7$      &  0.5  \\ 
                 & Occ.(1,1,9)           &  4          &    4    &   99&  3       &   148      & 149   &   0.03  &  0.3  &  $0.2\pm0.1$    &  0.2  \\
                 & Pa.([0,1],14,14,[1,9])     &  93         &    93   &   0.2        &  90        &    149     & 115&   33    &  0.3 &  $3\pm0.4$       &  0.5  \\  
                 & Pa.([0,1],1,1,[1,9])       &  7          &    8    &   71   &  4       &    154     & 140&   0.6   &  0.4  &  $4\pm0.5$      &  0.5  \\ 
                 & Pa.([0,1],1,1,[1,5])       &  2          &    3    &   131  &  2       &    151     & 150&   0.1   &  0.2  &  $0.3\pm0.1$    &  0.5  \\ 
\midrule
MNIST            & $L_\infty$([0,0.1])    &    49     &   59    &   166  &     17     &       202     & 180 &    18   &  0.2    &  $4.3\pm0.6$  &   0.2  \\
$conv1$          & $L_\infty$([0,0.05])   &    23     &   44    &   180  &     3     & 181           &  180&    12   &  0.3    &  $3.4\pm1.0$  &   0.2 \\
$\delta^m=38.6$  & Brightness([0,0.25])   &    29     &   31    &   94   &     4    &   300      & 180 &    6    &  0.6    &    $1.1\pm0.7$&   0.2  \\ 
$\delta^d=30$\%  & Brightness([0,0.1])    &    15     &   22    &   180  &     5    &   233      & 180 &    4    &  0.8    &    $1.0\pm0.5$&   0.3  \\
                 & Occ.(14,14,9)           &    36     &   36    &   1.2  &     10   &   195      & 180 &    9    &  0.5    &   $1.7\pm0.5$ &    0.3\\ 
                 & Occ.([12,16],[12,16],5) &    37      &   37   &   3.3  &     21   &   307      & 180&    10    & 2.5     &   $1.2\pm2.4$ &    0.9 \\
                 & Occ.(1,1,5)             &    9      &   9     &   0.8  &     4    &   300      & 180&    0    & 0.5     &   $0.4\pm1.8$ &    0.2 \\  
                 & Pa.([0,1],14,14,[1,9]) &    36     &   36    &   0.5   &     17   &   191     & 180 &    13   &  0.4    &    $3.6\pm0.5$&   0.5 \\  
                 & Pa.([0,1],14,14,[1,5]) &    26     &   27    &   0.4  &     5    &   212      & 180 &    12   &  0.4    &    $2.1\pm0.7$&   0.5 \\ 
                 & Pa.([0,1],14,14,[1,3]) &    13     &   13    &   0.8  &     24   &   187      & 180 &    6    &  0.4    &    $1.3\pm1.0$&    0.5\\ 
                 & Trans.([1,3],[1,3])    &    93     &   93    &   3.8  &     5    &   212      & 180&    19   & 1.6     &   $2.5\pm1.5$ &   1.1  \\ 
                 & Rotation($10^\circ$)   &    80     &   81    &   1.6  &     33   &   155      & 180 &    11   &  1.5    &  $1.5\pm0.6$  &   1.3 \\                  
\midrule
MNIST            & $L_\infty$([0,0.05])       &    25     &   25    &   4.9  &     0      &   304          &   180  &   12    &  0.3  & $1.4\pm0.02$  &  0.3  \\
$conv2$          & Brightness([0,0.25])       &    21     &   23    &   65  &   1           &      373       &  180   &   8     &  0.5  & $0.4\pm0.03$  &  0.3  \\ 
$\delta^m=147$   & Brightness([0,0.1])        &    9      &   13    &   88  &    7          &    328         & 180    &   5     &  0.7  & $0.4\pm0.03$  &  0.3  \\
$\delta^d=42$\%  & Occ.([12,16],[12,16],5)     &    37     &   38    &   4.9  &    0         &   321          & 180    &   17     &  3.3  & $2.5\pm1.0$  &  1.6 \\
                 & Occ.(14,14,5)               &    18     &   19    &   2.4  &    0         &  307           & 180    &   5     &  1.3  & $0.3\pm0.07$  &  0.2 \\
                 & Occ.(1,1,5)                 &    2      &   2     &   1.8  &     1        &     303        & 180    &   0     &  1.2  & $0.1\pm0.1$ &  0.3 \\
                 & Pa.([0,1],14,14,[1,9])     &    61     &   61    &   0.7 &    1          &     303        & 180     &   14    &  0.4  & $0.9\pm0.06$  &  0.4 \\  
                 & Pa.([0,1],14,14,[1,5])     &    27     &   28    &   2.7 &     0         &    301         & 180    &   11    &  0.5  & $0.8\pm0.05$  &  0.4  \\ 
                 & Pa.([0,1],14,14,[1,3])     &    13     &   13    &   3   &    0          &      295       & 180    &   7     &  0.4  & $0.6\pm0.07$  &  0.4  \\  
                 & Trans.([1,3],[1,3])        &    99     &   99   &   3.5  &     0         &      302       &  180   &   25    &  1.1  & $1.0\pm0.02$  &  1.2 \\ 
                 & Rotation($10^\circ$)       &    79     &   79    &   1.6  &     0        &     595        &  180   &   15    &  1.0  &  $0.7\pm0.05$ &  0.9  \\                  
\bottomrule        
\end{tabular}
    \label{tab:MNIST_tests}
\quad
\end{center}
\end{table*}

\begin{table*}[t]
\small
\begin{center}
\caption{\tool vs. Marabou and sampling approaches over Fashion-MNIST and CIFAR-10 classifiers. Sa. abbreviates Sampling, Occ. Occlusion, Pa. Patch, and Trans. Translation.}
\begin{tabular}{ll ccc cc cc ccc}
\toprule
Model & Perturbation             & \multicolumn{3}{c}{\tool} & \multicolumn{3}{c}{Marabou} & \multicolumn{2}{c}{ Dataset Sa.} & \multicolumn{2}{c}{Random Sa.}\\
\cmidrule(lr){3-5} \cmidrule(lr){6-8} \cmidrule(lr){9-10}  \cmidrule(lr){11-12} 
      &                          & $\delta^l$ &  $\delta^u$ &  $t$ & $\delta^l$ &  $\delta^u$ &  $t$ & $\delta^{DS}$ & $t$ & $\delta^{RS}\pm h$  & $t$ \\
      &                          & \%         &  \%         &   $[m]$     & \%         &  \%         &  $[m]$ & \%         &    $[m]$   & \%          & $[m]$ \\
\midrule
FMNIST           & $L_\infty$([0,0.05])          &    30     &    59   &   178 &   7       &   224       & 180   &   16     &  0.3  & $6.1\pm0.5$  &    0.5  \\
$conv1$          & Brightness([0,0.1])          &    17     &    52   &   118  &   14      &   130      &  180 &   2     &  0.4  & $0.5\pm0.1$   &    0.6   \\
$\delta^m=66.8$  & Occ.(14,14,9)               &    32     &    33   &   19   &  28       &    154     & 180&   5     &  0.6  & $0.4\pm0.2$   &    0.4    \\ 
$\delta^d=46$\%  & Occ.(1,1,5)                 &    17     &    17   &   2.9  &  15      &    151     & 180 &   0.2   &  0.5  & $0.3\pm0.15$  &    0.4   \\  
                 & Pa.([0,1],14,14,[1,9])     &    52     &    53   &   12.5 &    42     &    151     & 180&   7     &  0.2  & $0.8\pm0.1$   &    0.5  \\  
                 & Pa.([0,1],14,14,[1,5])    &    19     &    19   &   5.2  &   15      &    151     & 180 &   4     &  0.2  & $0.7\pm0.1$   &    0.5   \\ 
                 & Pa.([0,1],14,14,[1,3])    &    11     &    11   &   9.2  &   7       &   153     & 180  &   2     &  0.2  & $0.3\pm0.1$   &    0.5   \\ 
                 & Trans.([1,3],[1,3])            &    92     &    94   &   16.3 &  90      &    151     & 180 &   17    &  1.1  & $0.9\pm0.15$  &    0.7   \\ 
                 & Rotation($10^\circ$)       &    89     &    89   &   2.1  & 82       &    151     & 180&   11    &  1.6  & $1.05\pm0.2$  &   1.1     \\                  
\midrule
CIFAR-10         & Occ.(1,1,9)                 &    24     &    25   &  7.4  &    1       & 308            & 180    &   6     &  0.3  & $0.6\pm0.2$   &  0.3      \\
$conv1$          & Occ.(1,1,5)                 &    15     &    18   &  26   &      0     &     300        & 180    &   3     &  0.3  & $0.4\pm0.3$   &  0.2     \\    
$\delta^m=54.3$  & Pa.([0,1],14,14,[1,9])     &    51     &    62   &  63    &    1        &     313        & 180   &   8     &  0.5  & $1.3\pm0.1$   &  0.6      \\  
$\delta^d=43$\%  & Pa.([0,1],14,14,[1,5])     &    29     &    34    &  35    &    0       &    303        &   180  &   3     &  0.5  & $0.7\pm0.2$   &  0.5     \\ 
                 & Pa.([0,1],14,14,[1,3])     &    14     &    22    &  51    &         0  &       307      &  180  &   1     &  0.5  & $0.4\pm0.3$   &  0.5      \\ 
             
\bottomrule        
\end{tabular}
    \label{tab:FMNIST_CIFAR10_tests}
\quad
\end{center}
\end{table*} 

\paragraph{Comparison to baselines} 
We begin by evaluating all approaches for various perturbations: $L_{\infty}$, brightness, contrast, occlusion, patch, translation, and rotation. 
For each perturbation, we run each approach for every network and every pair of a class $c'$ and a target class (i.e., $C_t=\{c_t\}$) for three hours. 
 We record the execution time in minutes $t$, the lower bound $\delta^l_{c'}$ (of all approaches) and the upper bound $\delta^u_{c'}$ (of \tool and Marabou). 
 To put these values in context, we 
 report the average percentages, ${\delta^l_{c'}}/{\delta^m_{c'}}\cdot 100$ and ${\delta^u_{c'}}/{\delta^m_{c'}}\cdot 100$, where
 $\delta^m_{c'}$ 
 is the maximum class confidence of $c'$ 
 (computed by solving $\max\;\mathcal{C}(x,c',D) \texttt{ subject to } \mathcal{D}$). 
 We further report $\delta^d_{c'}\triangleq{\delta'_{c'}}/{\delta^m_{c'}}\cdot 100$, where ${\delta'_{c'}}$ is the maximum class 
 confidence of $c'$ in the given dataset. 
 Although global robustness provides guarantees for any input, including those outside the dataset, 
this value provides context to how close (and relevant) the minimal globally robust bound is to dataset-like inputs. 
\Cref{fig::demonstration} provides a visualization of the bounds computed by all approaches, while
\Cref{tab:MNIST_tests} provides the results for MNIST and \Cref{tab:FMNIST_CIFAR10_tests} for Fashion-MNIST and CIFAR-10 (\ifthenelse{\boolean{is_conference}}{we provide more results in the extended version of this paper~\cite[Appendix B]{g_ref_81}}{\Cref{sec:appex_eval} provides more results}). 
We note that if a perturbation's parameter (e.g., a coordinate or an angle) is over $[v,v]$, we denote it as $v$.
Our results show that (1)~the average gap between the lower and upper bound of \tool is 1.9, while Marabou's gap is 154.7, (2)~the lower bound of \tool is greater (i.e., tighter) by 6.5x than Marabou, by 9.2x than Dataset Sampling and by 25.6x than Random Sampling, (3)~the upper bound of \tool is smaller
 (i.e., tighter) by 13.1x than Marabou, (4)~\tool is 130.6x faster than Marabou, and (5) \tool is 115.9x slower than the sampling approaches 
(which is not surprising since these provide very loose lower bounds). 
 The ratio of \tool's $\delta^u$ and $\delta^m$ is 0.37 on average.  
Additionally, although \tool considers any possible input, the results show that our $\delta$-globally robust bounds overlap with the class 
confidences of the dataset's inputs: the ratio of \tool's $\delta^u$ and $\delta^d$ is between 0.02-3.1 and on average 0.9. 
We note that without the rotation and translation, whose bounds are relatively high, the average is 0.69. 
The high bounds of small rotation and translation perturbations are counterintuitive, 
since these perturbations are imperceptible. These counterintuitive bounds highlight the importance of computing the minimum globally robust bounds to gain insights into the network's global robustness against perturbations. 
In \ifthenelse{\boolean{is_conference}}{the extended version of this paper}{~\Cref{sec:appex_eval}}, we show the execution time of each component of \tool.

 \begin{figure}[t]
    \centering
  \includegraphics[width=1\linewidth, trim=0 114 180 0, clip,page=2]{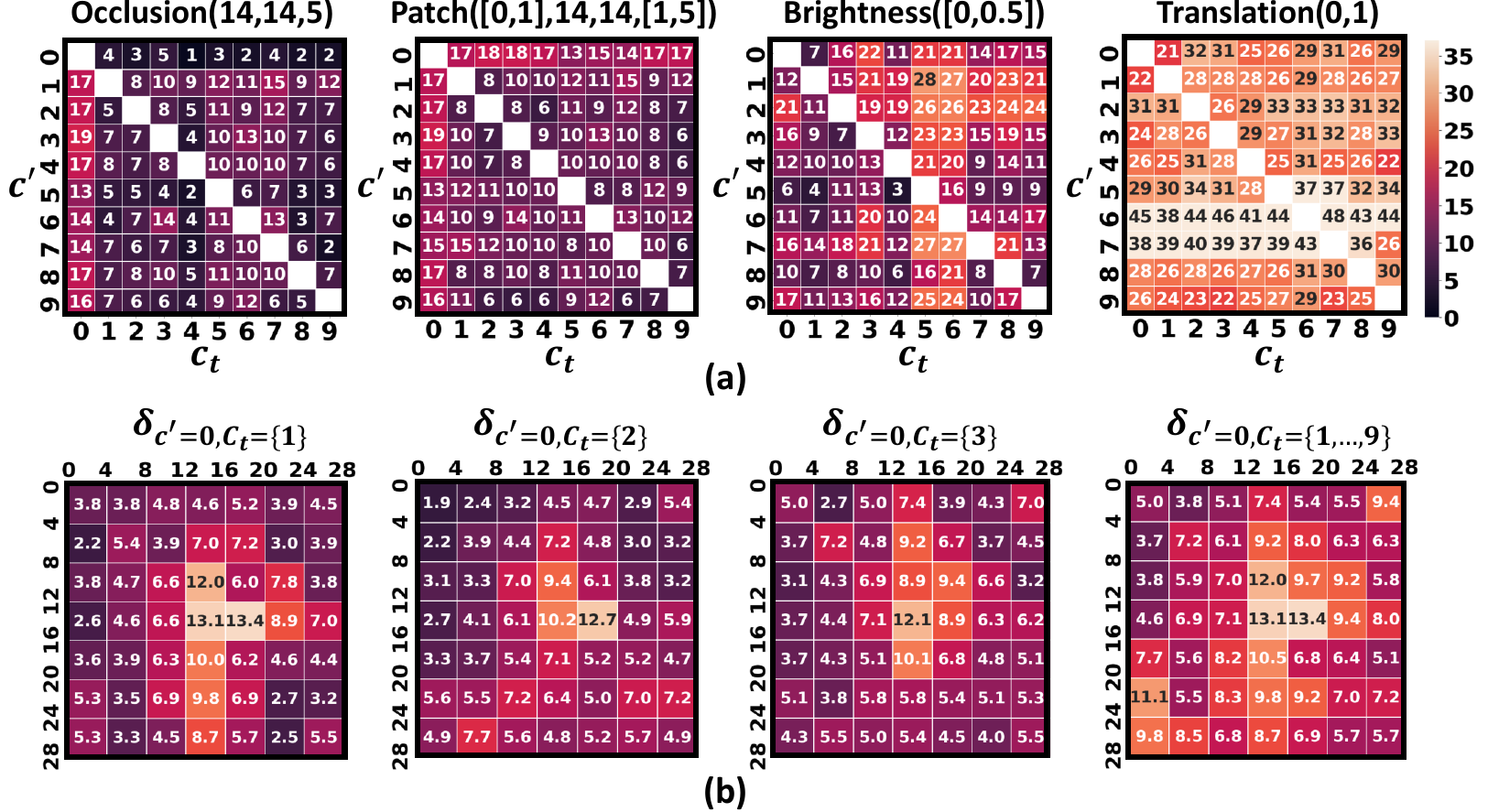}
    \caption{Visualization of the minimal globally robust bounds over various perturbations, for MNIST \texttt{conv1}.}
    \label{fig::visualization_and_comparing_the_attacks}
\end{figure}

\paragraph{Interpreting robustness bounds}
We next show how our minimal globally robust bounds can reveal global robustness and vulnerability attributes of a network to an adversarial attack.
We consider two experiments, both focus on the \texttt{conv1} network for MNIST. 
In the first experiment, we compare the network's global robustness bounds for four perturbations. 
For each perturbation, \tool computes the minimal globally robust bound for every pair of a class $c'$ and a target class $C_t=\{c_t\}$. 
\Cref{fig::visualization_and_comparing_the_attacks}(a) shows the bounds in a heatmap: the darker the entry
the lower the minimal globally robust bound is (i.e., the harder it is to attack). 
The heatmaps demonstrate the following. 
First, there is a significant variance in the bounds of different target classes, e.g., 
 $\delta_{c'=5,c_t=0}=13$ while $\delta_{c'=5,c_t=4}=2$, for occlusion(14,14,5). 
Namely, any image whose class confidence in $c'=5$ is more than 2 cannot be classified as $c_t=4$ by this attack. Among these images, only those whose class confidence is more than 13 are globally robust to $c_t=0$ for this attack. That is, it is significantly harder for an attacker to use this occlusion attack in order to fool \texttt{conv1} into classifying a five-digit image as a four, compared to as a zero.
Second, there is a variance in the effectiveness of the perturbation attacks. 
As expected, the bounds of the occlusion(14,14,5) attack are smaller or equal to the bounds of the patch([0,1],14,14,[1,5]) attack, which 
subsumes it. 
Also, generally, the classifier is less globally robust to the brightness([0,0.5]) attack compared to the occlusion(14,14,5) and the patch([0,1],14,14,[1,5]) attacks. Surprisingly, even though the translation(0,1) attack may seem to introduce a small perturbation, the classifier is not very robust to it (i.e., the bounds are very high).     
By such inspection, \tool enables the network designer to understand what perturbations are effective or not. 
The second experiment looks for spatial attributes of global robustness. 
In this experiment, we partition pixels into $4\times 4$ squares and the goal is to identify which squares are more robust to arbitrary perturbations, for inputs classified as $c'=0$. This can be determined by running \tool with the patch perturbation for every square with $\epsilon=1$. 
\Cref{fig::visualization_and_comparing_the_attacks}(b) shows heatmaps over the input dimension, where each entry shows the bound returned by \tool for $C_t=\{c_t\}$, for $c_t\in\{1,2,3\}$, and $C_t=\{1,\ldots,9\}$ (i.e., an untargeted attack).    
The heatmaps show that \texttt{conv1} is more vulnerable to patch attacks at the center, where the digit is located. 
However, an untargeted attack can also mislead the network by perturbing a patch in the background. These heatmaps allow the network designer to identify vulnerable pixel regions and consider a suitable defense.

\begin{table*}[t]
\small
\begin{center}
\caption{Ablation study. The first two rows are over MNIST $3\times 50$; the other rows are 
over MNIST $conv1$. }
\begin{tabular}{lccccccccccccccc}
\toprule
Perturbation             & \multicolumn{5}{c}{\tool} & \multicolumn{5}{c}{ Our MIP + dependencies} & \multicolumn{5}{c}{Our MIP} \\
\cmidrule(lr){2-6} \cmidrule(lr){7-11} \cmidrule(lr){12-16} 

                     & $t$ & $t^i$ & $\delta^l$ & $\delta^u$ & $\delta^i$ & $t$ & $t^i$ & $\delta^l$ & $\delta^u$ & $\delta^i$ & $t$ &  $t^i$ & $\delta^l$ & $\delta^u$ & $\delta^i$   \\
                     & $[m]$ & $[m]$ & $\%$ & $\%$ & $\%$& $[m]$ & $[m]$ & $\%$ &$\%$  &  $\%$& $[m]$ & $[m]$ & $\%$ & $\%$ & $\%$   \\
\midrule
Brightness([0,1])          & 4.8 & 0.1 & 48 & 48 & 39        & 8.2 & 1.2 & 48 &  48 & 30 &     18.5 & 1 & 48 & 48 & 27 \\
Brightness([0,0.1])        & 9.2 & 0.2 & 21 & 21 & 15        & 18 & 1 & 21 &  21 & 12 &     34.6 & 1.1 & 21 & 21 & 15 \\
\midrule
Brightness([0,1])          & 44 & 0.7 & 44 & 44 & 36            & 53  & 6.5 & 44  & 44   & 31  &     91 & 12 & 44 & 49 & 23 \\
Occ.(14,14,3)             & 0.9  & 0.15 & 9 & 9 & 7      & 0.9 & 0.3 & 9 &  9 & 6 &     180 & 40 & 4 &  24 & 3 \\
Pa.([0,1],14,14,[1,3])     & 0.7  & 0.05 & 13 & 13 & 6      & 0.9 & 0.1 & 13 &  13 & 5 &     180 & 0.1 & 13 &  34 & 4 \\
Translation(3,3)         & 0.1   & 0.01 & 93 & 93 & 90           & 0.3 & 0.1 & 93 &  93 & 67 &     0.5 & 0.1 & 93 &  93 & 66 \\

\bottomrule
\end{tabular}
    \label{tab:ablation}
\quad
\end{center}
\end{table*} 
\paragraph{Ablation study}  Lastly, we study the importance of \tool's components via an ablation study. We compare \tool to two variants: MIP-only and MIP + dependencies (i.e., without the hyper-adversarial attack). 
 We run all approaches over the MNIST $3\times 50$ and $conv1$ networks and various perturbation types, for every pair of classes $c'$ and $c_t$, for three hours.
 We measure the execution time in minutes $t$, the upper bound $\delta^u$, and the lower bound $\delta^l$.
  We also measure the minute at which the first non-trivial lower bound is obtained $t^i$ and its corresponding class confidence $\delta^i$. 
  Note that the first non-trivial lower bound depends on the variant's MIP (i.e., with/without dependencies, with/without the hyper-adversarial attack's lower bound). 
  The higher the $\delta^i$ the closer the optimization is to the maximal globally non-robust bound, while the shorter the $t^i$ the more efficient the variant is. 
 \Cref{tab:ablation} shows the results. Our results indicate that \tool is 1.7x faster than the MIP + dependencies variant and 78.6x faster than the MIP-only variant. 
 Given three hours, the upper and lower bounds of \tool are identical to the MIP + dependencies variant. The lower bound of the MIP-only variant is smaller (i.e., looser) by 1.3x and the upper bound is greater (i.e., looser) by 1.05x. 
 Also, the results show that \Cref{alg:sub_opt} allows \tool to obtain 
 1.2x (1.5x) higher (i.e., tighter) initial solutions in 6.7x (52.6x) less time compared to the MIP + dependencies variant (MIP-only variant).

\section{Related Work}
\label{sec:related_work}


\paragraph{Global robustness} Several works analyze global robustness properties. \citet{g_ref_30,g_ref_1,g_ref_2} compute the worst-case change in a network's outputs but cannot determine whether the classification changes. 
\citet{g_ref_41,g_ref_42} estimate global robustness for classifiers from local robustness guarantees. \citet{g_ref_43,g_ref_44} provide probabilistic guarantees. 
\citet{g_ref_80} compute robustness guarantees for input regions derived from the dataset; however, these do not imply global robustness for any input.
 \citet{g_ref_45,g_ref_77} train $L_\infty$ globally robust networks using Lipschitz bounds.
  \citet{g_ref_78} are similar but focus on SDN (and not DNN). \citet{g_ref_47} propose booster-fixer training for security classifiers to enforce global robustness.  

\paragraph{Multiple network encodings} 
\tool encodes two copies of a network classifier to analyze the output of an input and its perturbed example. 
Several works employed a similar concept. \citet{g_ref_30,g_ref_1,g_ref_2} analyze global robustness by reasoning about two copies of a network (not a classifier) and bounding the maximum change in the output caused by an input perturbation. \citet{g_ref_1,g_ref_2} leverage dependencies defined by interleaving connections which capture the difference of respective neurons to overapproximate their functionality. In contrast, \tool integrates dependencies stemming from the perturbation, propagates dependencies across layers, and infers neurons' dependencies from their concrete bounds or by solving suitable MIP problems. 
Other works reason about multiple networks in the context of differential analysis~\citep{g_ref_60} or local robustness of an ensemble of networks~\citep{g_ref_61}.
  
\paragraph{Attack-guided verification} \tool executes a hyper-adversarial attack to obtain suboptimal feasible solutions. Several works integrate adversarial attacks with verification. \citet{g_ref_22} improve a verifier's precision by leveraging spurious adversarial examples. 
 \citet{g_ref_72} employ attacks to train networks with provable robustness guarantees. \citet{g_ref_75} leverage the PGD attack~\citep{g_ref_6} to compute robust adversarial regions.

\section{Conclusion}
\label{sec:conclusions_and_discussion}
We present \tool for computing the minimal globally robust bound of a neural network classifier, under a given perturbation.
 \tool encodes the problem as 
 a MIP, over two network copies, which is solved in an anytime manner. 
 \tool reduces the MIP's complexity by (1)~identifying dependencies stemming from the perturbation and the network's computation and (2)~generalizing adversarial attacks to unknown inputs to 
 compute suboptimal lower bounds and 
 optimization hints. 
Our results show that \tool computes 82.2x tighter bounds and is 130.6x faster than an existing global robustness verifier. 
Additionally, its lower bounds are 18.8x higher than those computed by sampling approaches. 
Finally, we exemplify how \tool can provide insights into the robustness attributes of a network classifier to adversarial attacks.  

\section*{Acknowledgements} We thank the anonymous reviewers for their feedback. This research was supported by the Israel Science Foundation (grant No. 2605/20). 

\section*{Data-Availability Statement}
Our code is available at \url{https://github.com/ananmkabaha/VHAGaR.git}. 
\bibliography{bib}
\newpage
\appendix
\ifthenelse{\boolean{is_conference}}{}{\appendix
\section{Image Rotation with Bilinear Interpolation}\label{sec:appex_rot}
\Cref{alg:img_rot} shows the standard rotation algorithm that \tool relies on.

\begin{algorithm}[t]
\DontPrintSemicolon
\KwIn{An input image $x\in[0,1]^{d_1\times d_2}$ and a rotation angle $\theta \in [0,2\pi]$. }
\KwOut{The rotated image $x_R\in[0,1]^{d_1\times d_2}$.}
$center = [d_1/2+1,d_2/2+1]$\;
\For {$i\in \{1,\dots,d_1\}$, $j \in \{1,\dots,d_2\}$ }{
    $i_c=i-center[1]$\;
    $j_c=j-center[2]$\;
    $i_r = (i_c \cdot \sin(\frac{\theta \cdot \pi}{180}) + j_c \cdot \cos(\frac{\theta \cdot \pi}{180})) + center[1]$\;
    $j_r = (i_c \cdot \cos(\frac{\theta \cdot \pi}{180}) - j_c \sin(\frac{\theta \cdot \pi}{180}) + center[2]$\;
    \If {$\lfloor i_r \rfloor \geq 1$ and $\lceil i_r \rceil \leq d_1$ and $\lfloor j_r \rfloor \geq 1$ and $\lceil j_r \rceil \leq d_2$}{
        $di = i_r - \lfloor i_r \rfloor$\;
        $dj = j_r - \lfloor j_r \rfloor$\;
        $x_R[i, j] = (1 - di) \cdot (1 - dj) \cdot x[\lfloor i_r \rfloor, \lfloor j_r \rfloor] + di \cdot (1 - dj) \cdot x[\lceil i_r \rceil, \lfloor j_r \rfloor] + (1 - di) \cdot dj \cdot x[\lfloor i_r \rfloor, \lceil j_r \rceil] + di \cdot dj \cdot x[\lceil i_r \rceil,\lceil j_r \rceil]$    
    }    
}
\Return{$x_R$}
  \caption{Image\_Rotation\_with\_Bilinear\_Interpolation($x$, $\theta$)}\label{alg:img_rot}
\end{algorithm}

\section{Additional Results}\label{sec:appex_eval}
\Cref{tab:add_res}
provides additional results comparing \tool to the baselines. This table also shows a $3\times 50$ model for MNIST trained without PGD, denoted $3\times 50$-ND (for no defense).
\Cref{fig::Time split} shows the execution time of \tool's components: the concrete bound computation, the computation of the dependency constraint, the hyper-adversarial attack, and the time to solve the main MIP (Problem~\ref{main_problem}). 
The pie charts show the execution times of the different components as if they run sequentially, however \tool runs the hyper-adversarial attack in parallel to the concrete bound computation and the dependency constraint computation. Additionally, the concrete bound computation and the dependency constraint computation are run once and not separately for each $c_t\in C_t$. Results show that the dominant factor of \tool's execution time is solving Problem~\ref{main_problem}, while the other computations terminate within two minutes combined.

\begin{center}
\begin{small}
\begin{longtable}{ll ccc cc cc ccc}
\caption{Additional results: \tool vs. sampling baselines and Marabou. Sa. abbreviates Sampling, Occ. Occlusion, Pa. Patch, and Trans. Translation. NA is not applicable.}\label{tab:add_res}\\
\toprule
Model & Perturbation             & \multicolumn{3}{c}{\tool}& \multicolumn{3}{c}{Marabou} & \multicolumn{2}{c}{ Dataset Sa.} & \multicolumn{2}{c}{Random Sa.} \\
\cmidrule(lr){3-5} \cmidrule(lr){6-8} \cmidrule(lr){9-10}  \cmidrule(lr){11-12} 
      &                          & $\delta^l$ &  $\delta^u$ &  $t$ & $\delta^l$ &  $\delta^u$ &  $t$ & $\delta^{DS}$ & $t$ & $\delta^{RS}\pm h$  & $t$ \\
      &                          & \%         &  \%         &   $[m]$ & \%         &  \%         &  $[m]$    & \%         &    $[m]$   & \%          & $[m]$  \\
\midrule
MNIST             & $L_\infty$([0,0.5])    &  99         &    99   &   0.1  &    87    &   154         &  180 &   57    &  0.2  &  $7.5\pm5.4$    &  0.2  \\
$3\times 50$              & Brightness([0,1])      &  48         &    49   &   4.8  &  31      &    151     & 131 &   17    &  0.4  &  $4\pm0.6$      &  0.4  \\ 
 $\delta^m=3.3$               & Occ.(14,14,5)         &  60         &    60   &   1.6  &  51      &    154     & 139  &   16    &  0.2  &  $2\pm0.6$      &  0.2  \\
$\delta^d=73$\%   & Occ.(14,14,3)         &  36         &    36   &   5.4   &  33      &    160     & 130 &   7     &  0.5  &  $2\pm0.6$      &  0.2 \\
   & Occ.(1,1,5)           &  0.6        &    2    &   156  &  2       &   152      & 141 &   0     &  0.3  &  $0.03\pm0.1$   &  0.2  \\ 
                  & Pa.([0,1],14,14,[1,5])     &  85         &    85   &   0.5  &  78      &    160     & 118&   32    &  0.4  &  $3\pm0.4$      &  0.5 \\ 
                  & Pa.([0,1],14,14,[1,3])     &  61         &    62   &   2.1  &  54      &    157     & 121&   26    &  0.4  &  $5\pm0.4$      &  0.5  \\ 
\midrule
MNIST              & Brightness([0,0.25])   &    54     &   55    &   25    & 42         &  137         &  155 &   23    &  0.5  &    $2\pm16$   &  0.6  \\ 
 $3\times 50$-ND            & Brightness([0,0.1])    &    36     &   36    &   33    & 27         &  125       &  167 &   18    &  0.6  &    $1.7\pm16$ &  0.6  \\ 
$\delta^m=2.84$  & Contrast([1,2])        &     3     &   34    &   172    & NA         &  NA         & NA  &   0.5   &  0.6  &    $1.5\pm14$ &  0.6 \\
$\delta^d=45$\%  & Contrast([1,1.5])      &     3     &   30    &   180   & NA         &  NA         & NA  &   0.2   &  0.5  &    $1\pm15$   &  0.5  \\
\midrule
MNIST            & $L_\infty$([0,0.5])    &    98     &   98    &   1.2 &    1      &     207       & 180  &    24   &  0.3    &  $2.9\pm0.3$  &   0.2  \\
$conv1$          & Brightness([0,1])      &    43     &   44    &   44  &     22   &   155      & 180  &    8    &  0.8    &    $2.1\pm1.1$&   0.2 \\ 
$\delta^m=39$  & Occ.(14,14,5)         &    20     &   20    &   0.4 &     0.6  &   215      & 180 &    4    &  0.6    &   $0.8\pm1.1$ &    0.3 \\
$\delta^d=30$\%  & Occ.(14,14,3)         &    9      &   9     &   0.9 &     17   &   151      & 180  &    1    & 0.6     &   $0.5\pm1.1$ &    0.3\\
                 & Occ.(1,1,9)           &    24     &   25    &   0.2 &     6    &   233      & 180  &    1    & 0.6     &   $1.3\pm0.7$ &    0.2\\
                 & Pa.([0,1],1,1,[1,9])       &    36     &   37    &   0.3 &     2    &   211      & 180  &    11   &  0.4    &    $3.1\pm1.0$&    0.5\\ 
                 & Pa.([0,1],1,1,[1,5])       &    10     &   11    &   0.7 &     33   &   155      & 180  &    4    &  0.4    &    $1.1\pm1.0$&    0.5\\ 
                 & Trans.(0,[1,3])        &    90     &   90    &   1.5 &     21   &   155      & 180 &    15   &  0.4    &   $2.0\pm1.5$ &   0.4  \\ 
                 & Trans.([1,3],0)        &    90     &   92    &   1.8 &     17   &   191      & 180  &    19   & 0.5     &   $2.3\pm1.3$ &    0.4\\ 
                 & Rotation($1^\circ$)  &    71     &   74    &   1.4 &     24   &   187      & 180  &    8    & 1.6     &   $0.7\pm0.5$ &  1.2  \\ 
                 & Rotation($5^\circ$)  &    77     &   80    &   1.2  &     31   &   211      & 180  &    9    & 1.4     &   $0.9\pm0.6$ &   1.3\\ 
\midrule

MNIST            & $L_\infty$([0,0.1])      &    49     &   50    &   2.4  &     0      &   305          &   180  &   8    &  0.3  & $1.5\pm0.03$  &  0.2  \\
 $conv2$                 & Brightness([0,1])      &    36     &   38    &   55  &     3      &   595      &   180  &   11    &  0.6  & $0.5\pm0.03$  &  0.3  \\
$\delta^m=147$                  & Occ.(14,14,9)         &    40     &   40    &   1.1  &     0     &     310        & 180   &   10    &  1.4  & $0.4\pm0.06$  &  0.3 \\ 
$\delta^d=42$\%    & Occ.(14,14,3)         &    8      &   8     &   2.5 &     0     &        303     &  180   &   2     &  1.4  & $0.1\pm0.06$  &  0.3  \\
                  & Occ.(1,1,9)           &    28     &   29    &   1.2  &     0     &    324         & 180   &   3     &  1.3  & $0.6\pm0.06$  &  0.3  \\ 
                 & Pa.([0,1],1,1,[1,9])       &    45     &   45    &   0.9 &    0       &    301        & 180    &   12    &  0.4  & $0.7\pm0.02$  &  0.4  \\ 
                 & Pa.([0,1],1,1,[1,5])       &    3      &   3     &   1.6 &    0       &    304         &  180  &   1     &  0.5  & $0.3\pm0.03$  &  0.4  \\ 
                 & Trans.(0,[1,3])        &    90     &   91    &   0.8 &    0      &    303        &  180   &   19    &  0.3  & $1.1\pm0.02$  &  0.3  \\ 
                 & Trans.([1,3],0)        &    87     &   87    &   1  &     0      &   304          &  180    &   24    &  0.4  & $0.7\pm0.03$  &  0.4  \\ 
                 & Rotation($1^\circ$)  &    59     &   61    &   2.1  &   0       & 536          & 180 &   10    &  1.2  & $0.5\pm0.06$  &  0.9   \\ 
                 & Rotation($5^\circ$)  &    61     &   62    &   2.1 &      0     &  631          & 180    &   11    &  1.2  & $0.6\pm0.08$  &  1.0  \\ 
\midrule                
FMNIST           & Brightness([0,1])      &    47     &    64   &   180 &    37    &    225     &180  &   6     &  0.3  & $0.6\pm0.1$   &    0.6   \\ 
$conv1$           & Brightness([0,0.25])         &    32     &    56   &   162 &   26       &   195       & 180   &   3     &  0.3  & $0.55\pm0.1$  &    0.6  \\
$\delta^m=66.8$   & Occ.(14,14,5)         &    13     &    13   &   4.2 &  12      &    151     & 180 &   2     &  0.6  & $0.15\pm0.1$  &    0.4   \\
$\delta^d=46$\%   & Occ.(14,14,3)         &    7      &    7    &   4.8 &  6       &    153     & 180 &   0.5   &  0.6  & $0.07\pm0.1$  &    0.4   \\
                   & Occ.(1,1,5)           &    17     &    17   &   2.9  &  15      &    151     & 180 &   0.2   &  0.5  & $0.3\pm0.15$  &    0.4  \\ 
                 & Pa.([0,1],1,1,[1,9])       &    69     &    70   &   0.6 &   63     &    151     & 180 &   8     &  0.2  & $0.6\pm0.05$  &    0.5   \\ 
                 & Pa.([0,1],1,1,[1,5])       &    28     &    28   &   4.7 &  25      &    156     & 180   &   4     &  0.2  & $0.7\pm0.1$   &    0.5   \\ 
                 & Trans.(0,[1,3])        &    89     &    89   &   7.9  &  87      &    151     & 180 &   14    &  0.3  & $1.0\pm0.1$   &    0.2   \\ 
                 & Trans.([1,3],0)        &    91     &    92   &   8.6   &  84      &    156     & 180 &   12    &  0.5  & $0.65\pm0.2$  &    0.2   \\ 
                 & Rotation($1^\circ$)  &    80     &    81   &   5    &  70      &    154     & 180&   6     &  1.8  & $0.7\pm0.2$   &    1.2   \\ 
                 & Rotation($5^\circ$)  &    85     &    86   &   4.5 & 75       &    153     & 180  &   7     &  1.6  & $0.9\pm0.15$  &   1.2     \\ 
\midrule
CIFAR10          & Occ.(14,14,9)         &    31     &    48   &  85   &    0       &      317       &  180   &   7     &  0.3  & $0.7\pm0.3$   &  0.3     \\ 
$conv1$          & Occ.(14,14,5)         &    16     &    25   &  46   &      0     &      307       &  180   &   3     &  0.3  & $0.6\pm0.3$   &  0.3      \\
$\delta^m=54.3$  & Occ.(14,14,3)         &     9     &    16   &  54   &    0       &       301      &  180   &   1     &  0.2  & $0.4\pm0.3$   &  0.3      \\ 
$\delta^d=43$\%  & Pa.([0,1],1,1,[1,9]) &    42     &    43   &  13   &     1      &      301       &  180   &   6     &  0.5  & $1.0\pm0.1$   &  0.5     \\ 
                 & Pa.([0,1],1,1,[1,5]) &    25     &    27   &  24   &  0         &      313       &  180  &   2     &  0.5  & $0.6\pm0.2$   &  0.5      \\ 
                 \bottomrule
\end{longtable}
\end{small}
\end{center}

 \begin{figure}[h]
    \centering
  \includegraphics[width=1\linewidth, trim=0 236 10 0, clip,page=16]{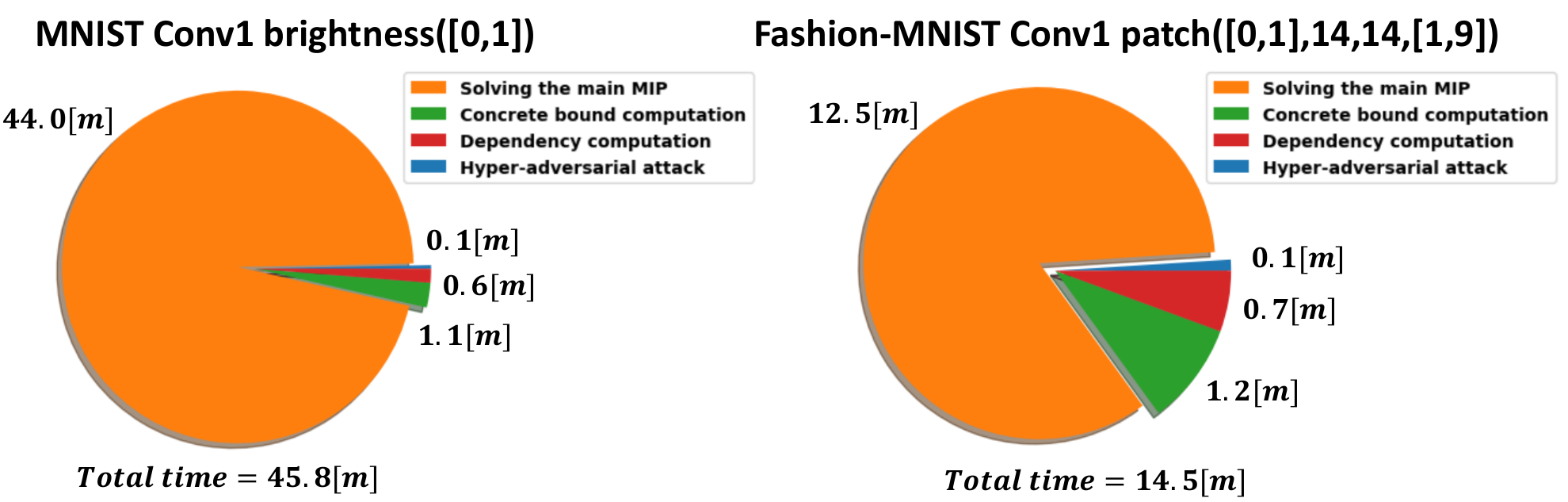}
    \caption{The execution time of \tool's different components.}
    \label{fig::Time split}
\end{figure}
}
\end{document}